\documentclass[11pt]{article}

\usepackage[letterpaper,margin=1in]{geometry}
\usepackage[utf8]{inputenc}
\usepackage[T1]{fontenc}
\usepackage{times}
\usepackage{natbib}
\bibpunct{(}{)}{;}{a}{,}{,}

\usepackage{amsmath,amssymb,amsthm}
\newtheorem{proposition}{Proposition}
\usepackage{mathtools}
\usepackage{bm}
\usepackage{graphicx}
\graphicspath{{./figure/}{./images/}}
\usepackage[dvipsnames,table]{xcolor}
\usepackage{booktabs}
\usepackage{array}
\usepackage{multirow}
\usepackage{multicol}
\usepackage{wrapfig}
\usepackage{tabularx}
\usepackage{xspace}
\usepackage{pifont}
\usepackage[font=small,labelfont=bf,labelsep=period]{caption}
\usepackage{url}
\usepackage{hyperref}
\hypersetup{colorlinks=true,linkcolor=black,citecolor=black,urlcolor=blue}

\usepackage{fancyhdr}
\newcommand{\pms}[1]{\,{\tiny$\pm$#1}}

\newcommand{\Zstate}{\mathcal{Z}}
\newcommand{\dinner}{d_{\text{inner}}}

\newcommand{\Cchunk}{C_{\text{chunk}}}

\title{SurgicalMamba: Dual-Path SSD with State Regramming\\
for Online Surgical Phase Recognition}

\author{
  Sukju Oh \quad Sukkyu Sun\thanks{Corresponding author.} \\
  Department of Computer Science and Artificial Intelligence \\
  Dongguk University, Seoul 04620, Republic of Korea \\
  \texttt{\{dhtjrwn119, sukkyu.sun\}@dgu.ac.kr}
}

\date{}

\begin{document}
\maketitle

\begin{abstract}
Online surgical phase recognition must commit to a prediction at every
frame of a procedure that runs for hours, from past frames alone and
at a per-frame cost that does not grow with elapsed length. Structured state-space
duality (SSD)~\cite{dao2024transformers} meets that constraint, but
only by having the scan see a per-head \emph{scalar} transition, which
fixes both where the state puts a frame and how fast it decays. The
same views recur through an operation, so repeated content is written
over itself and can afterwards be told apart only by age. How fast to
decay is left to the step, and when the past stops being useful has to
be inferred from a loss that never marks the moment. Procedures run
long and change little visually from frame to frame, leaving the step
with little to select on. Phases also vary widely in length, so no
fixed rate serves as a fallback.
We address the two with two mechanisms. \emph{State regramming}
rotates the carried state at each chunk boundary, by an amount the
chunk's content decides, so where a frame is written
also depends on what has passed since: two occurrences of the same
view are held apart when different phases intervene, which no decay
rate can achieve once both have aged. \emph{Intensity-modulated
stepping} increases the decay at the annotated phase transitions, so
the state empties quickly where a phase ends and slowly in between and
the decay itself can be set for the longest phase. Both leave SSD's $N$-semiseparable
structure and $O(d)$ per-frame cost intact.
Across seven public benchmarks \emph{SurgicalMamba} reaches
state-of-the-art online accuracy and phase-level Jaccard
($94.6\%/82.7\%$ on Cholec80, $89.5\%/68.9\%$ on AutoLaparo) at
$312.88$\,fps on a single GPU. Adding the rotation alone to a plain Mamba2
improves multi-query associative recall (MQAR) wherever the recurrent
state is the binding constraint, indicating that the mechanism is not
specific to surgical video.
Code: \url{https://github.com/sukjuoh/Surgical-Mamba}

\end{abstract}

\section{Introduction}
\label{sec:intro}

Online surgical phase recognition (SPR) identifies the current phase of
an ongoing procedure from streaming endoscopic video, and is the
temporal foundation for context-aware operating-room
systems~\cite{maierhein2017sds, garrow2021ml, hashimoto2018ai,
mascagni2022computer}. Unlike offline segmentation the setting is
causal: the model commits to a prediction at frame $t$ from frames
$1{:}t$ alone, at bounded per-frame
latency~\cite{twinanda2016endonet, jin2018svrcnet}. A procedure runs
30--90 minutes~\cite{twinanda2016endonet, wagner2023heichole}, so a
recognizer must reason over an hour-long horizon while holding
per-frame cost fixed. Transformer- and memory-based recognizers meet
the first demand at the expense of the second, since attention scope
or an external feature cache grows with elapsed length.

We therefore build on Mamba2's structured state-space duality
(SSD)~\cite{gu2023mamba, dao2024transformers}, whose chunked training
and recurrent inference compute the same function, so one set of
weights trains on tensor cores and streams at $O(d)$ per frame
regardless of elapsed length. That efficiency comes from reducing the
state transition to a per-head scalar, which fixes both where the
state puts a frame and how fast it decays, and neither can be
changed inside the transition without losing the matrix form. We
change instead the state the scan hands from one chunk to the next.

A scalar transition only rescales the state between steps; it never
moves content from one direction to another. Where a frame is written
is therefore decided by that frame alone and stays put for the rest of
the procedure. In surgery the same instruments and views return
throughout, so a scene that recurs is written on top of its earlier
occurrence, and the only thing left to separate the two is their
difference in age, which vanishes once both are old.
\emph{State regramming} ($Z$) rotates the carried state at each chunk
boundary, by an amount that chunk's content decides. Where a frame
sits then depends on what has passed since it was written: the same
clipping scene lands in a different part of the state when a
dissection has intervened, and a read-out can pick out one occurrence
without the other. The rotation leaves SSD's $N$-semiseparable
structure and $O(d)$ per-frame cost intact (\S\ref{sec:tstate}).

The same scalar also fixes how fast the state empties. The step varies
from frame to frame, but nothing tells it when the past has stopped
being useful, and that has to be inferred from a frame-level loss that
never marks the moment. The step is computed from the current
frame, and procedures run long and change little visually from frame to frame,
leaving it with little to select on. Nor is a fixed rate a fallback.
Phases are not comparable in length, dissection running for minutes
where clipping passes in seconds, so a decay fast enough to forget the
previous phase by the end of a short one keeps eroding a long phase
from within. \emph{Intensity-modulated stepping}
($\lambda$) increases the decay at the phase boundaries. They are
annotated, so we supervise $\lambda$ toward them and let it empty the
state quickly where a phase ends and slowly in between; the decay
itself can then be set for the longest phase, and the wider the spread of phase
durations, the more this recovers (Appendix~\ref{app:lambda}).

Both mechanisms sit in a dual-path SSD block: a slow path carries
state across clips under $\lambda$-warped stepping, while a fast path
resets at clip boundaries. Under strict online evaluation we reach
state-of-the-art accuracy and phase-level Jaccard on seven public SPR
benchmarks at $312.88$\,fps on a single GPU, and ablations isolate each
component. Outside surgery, adding the rotation to a plain Mamba2
improves multi-query associative recall (MQAR) wherever the recurrent
state is the binding constraint.

Our contributions are (i) \textbf{state regramming}, a chunk-granular,
exactly norm-preserving, input-conditioned rotation of the SSD carried
state, so that what a frame reads from a past one depends on the
interval between them and not only on the two frames, while provably
preserving $N$-semiseparability and $O(d)$ per-frame inference; (ii) \textbf{intensity-modulated
stepping}, a supervised continuous-time time-warp of the
discretization step that separates rate control from Mamba2's
unsupervised content selectivity; (iii) a \textbf{streaming-compatible
design} in which both mechanisms and the dual-path carry preserve
SSD's chunk structure end to end, holding per-frame inference at
$O(d)$; and (iv)
\textbf{state-of-the-art online SPR} on seven benchmarks under strict
causal evaluation at $312.88$\,fps.

\section{Related Work}
 
\subsection{Surgical phase recognition}
\label{sec:rel_spr}

Early recognizers paired CNN backbones with recurrent or convolutional
temporal models~\cite{jin2018svrcnet, jin2020mtrcnet,
czempiel2020tecno}, and transformer-, memory-, and retrieval-based
models then extended temporal scope~\cite{jin2021tmrnet,
gao2021transsvnet, liu2025lovit, liu2023skit, yang2024surgformer,
yue2023cmtnet} at a streaming cost: per-frame attention grows with
elapsed length or is capped by a sliding window~\cite{wang2021oadtr,
xu2021lstr}. Three recent recognizers frame the setting we target.
\citet{rivoir2024bnpitfalls} show that BatchNorm is the obstacle to
end-to-end training in this setting, and that with a backbone without
it, such as ConvNeXt~\cite{liu2022convnext}, single-stage training is
competitive;
DACAT~\cite{yang2024dacat} retrieves the most correlated past clip
from an unbounded cache, growing as $O(Td)$ and assuming the most
informative past is the most visually similar; and
MTTR-Net~\cite{huang2025mttrnet} closes the train/test feature gap
with multi-teacher cross-mimicking, at the cost of $K{+}1$ encoders in
a two-stage pipeline. SSMs have entered the field in several
forms, as temporal modules~\cite{cao2024srmamba, zhang2024sprmamba},
within hybrid temporal stacks~\cite{cmanet2025}, and as an endoscopic
foundation model~\cite{endomamba2025}, though the bidirectional
temporal scan of~\citet{cao2024srmamba} is non-causal and therefore
$O(Td)$ online.
In all of them the recurrence advances at a uniform rate, and where a
frame is written into the state is fixed by that frame alone. Neither
of the two handles we contribute, $\lambda$ (\S\ref{sec:slowpath}) and
$Z$ (\S\ref{sec:tstate}), is available within a causal $O(d)$
recurrence.

\subsection{Rotations in recurrent and state-space models}
\label{sec:rel_rot}

Norm-preserving recurrences have a long history. Orthogonal and
unitary RNNs constrain the recurrence \emph{weight} matrix against
exploding and vanishing gradients~\cite{arjovsky2016urnn,
mhammedi2017ornn, jing2017eunn, helfrich2018orthogonal,
lezcano2019cheap}, with a single learned matrix applied at every step;
the Cayley parameterization we use comes from that line, though the
rotated object and its purpose differ. Rotation is also not new inside
state-space models: complex-diagonal transitions give each channel
pair a fixed rotation in its own plane~\cite{gu2022s4,
orvieto2023lru}, and Mamba-3~\cite{lahoti2026mamba3} reintroduces
complex transitions into SSD by computing them as data-dependent
rotary embeddings on the $B,C$ \emph{projections}.

What none of these rotate is the carried state itself. Constraining
the recurrence weights fixes how the state evolves but leaves its
contents in place, and acting on the projections changes how the state
is written and read without moving what is already stored. State
regramming rotates the stored state directly at chunk boundaries,
densely across all $N$ dimensions and conditioned on that chunk's
content; Appendix~\ref{app:mamba3} works out the resulting difference
from projection-side rotations in the SSD matrix form.

\subsection{Forgetting in recurrent sequence models}
\label{sec:rel_stepping}

Selective SSMs already decide how much to forget at every frame. In
Mamba the step $\Delta$ is computed from the current token and fixes
how much of the state survives it~\cite{gu2023mamba,
dao2024transformers}. Recent work has found two things wrong with how
that decision is made. Gated DeltaNet~\cite{yang2025gateddeltanet}
takes it to be too coarse, since a single $\Delta$ decays every
key--value association by the same factor, and adds a delta-rule term
so that a write can target one association without disturbing the
rest. \citet{chen2024stuffed} find instead that the decision is barely
learned at all, Mamba-2 failing to forget earlier content because
training on contexts short relative to the state size lets it do well
without ever needing to; \citet{wang2025memmamba} attribute the same
difficulty to exponential decay itself.

In all of these the model has to work out for itself when the past
stops being useful. Where a task marks that point, it need not.
Surgical video annotates its phase boundaries, and $\lambda$ is
supervised toward them (\S\ref{sec:loss}), so what is released and
when follows the structure of the procedure rather than whatever the
frame-level loss settles on. The price is that a task must expose such
boundaries at all.

\section{Method}
\label{sec:method}
 
\subsection{Preliminaries and notation}
\label{sec:preliminaries}

Mamba2~\cite{gu2023mamba, dao2024transformers} discretizes a
continuous-time selective SSM with input-dependent step
$\Delta_n=\Delta(x_n)$, giving the per-head recurrence
\begin{equation}
h_n=\bar A_n\,h_{n-1}+\bar B_n\,x_n,\qquad y_n=C_n\,h_n,
\qquad \bar A_n=\exp(\Delta_n A),\ \ \bar B_n\approx\Delta_n B_n,
\label{eq:ssd_recur}
\end{equation}
with $A<0$ a \emph{per-head scalar}, so $\bar A_n$ acts on the state
$h\in\mathbb{R}^{H\times P\times N}$ by element-wise decay. The
projections $B_n$ and $C_n$ are likewise computed from the frame, and
fix the two things the recurrence does with a direction of the state:
$B_n$ says where frame $n$ is written, and $C_n$ which directions the
output at $n$ reads. The SSD form computes \eqref{eq:ssd_recur}
chunk-wise: a clip of $L$ frames is split into $n_c=L/\Cchunk$ chunks,
each computed as one fused matrix multiply, with exactly one state
tensor passed across each chunk boundary. Dually, the same map is an
$N$-semiseparable matrix~\cite{dao2024transformers}, the property that
yields $O(d)$ per-frame inference and that our structural result
(Proposition~\ref{prop:preserve}) preserves. The zero-order-hold
derivation and the notation used throughout are in
Appendix~\ref{app:prelim}.

\subsection{Overview}
\label{sec:overview}

\begin{figure}[t]
\centering
\includegraphics[width=0.9\linewidth]{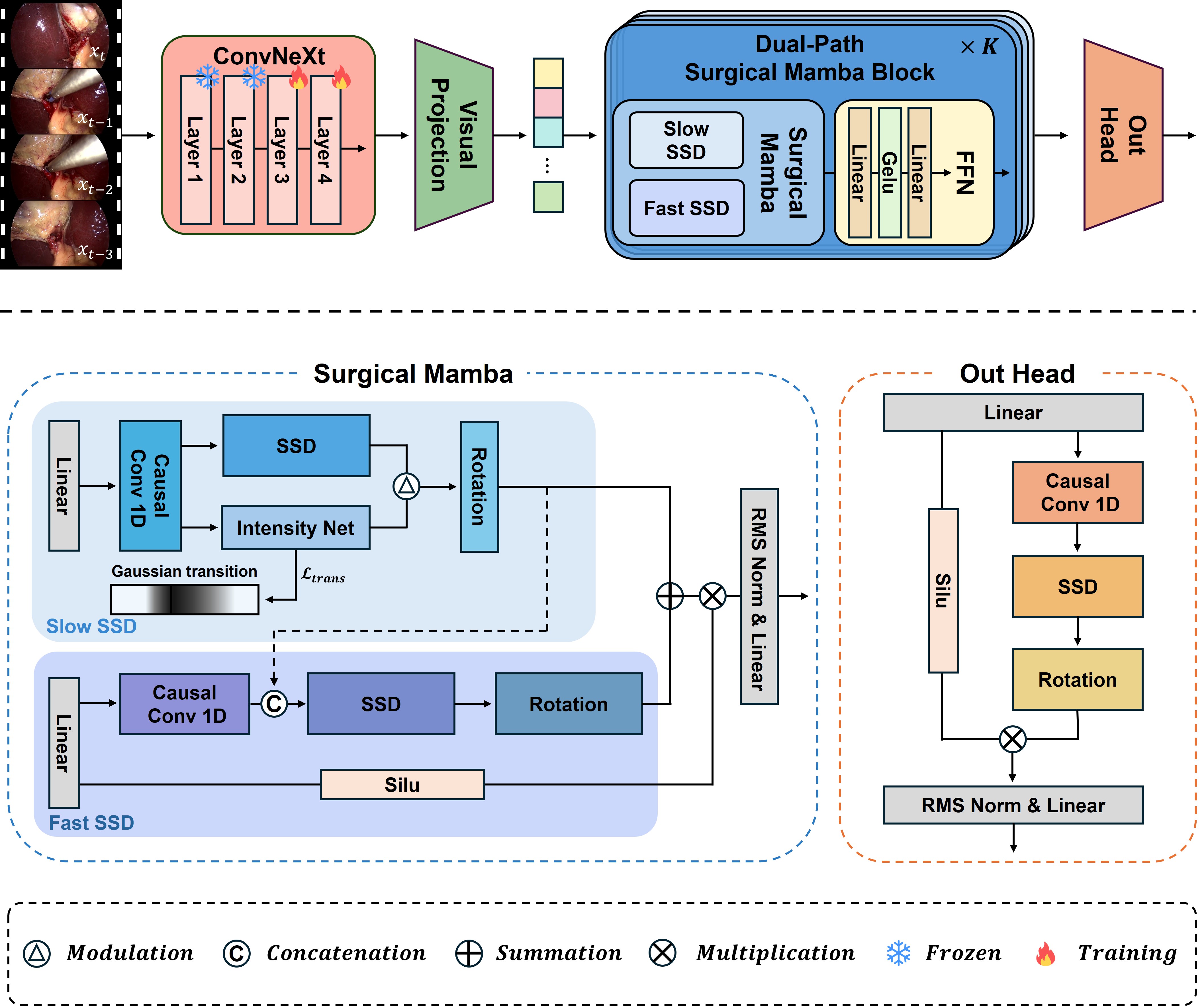}
\caption{\textbf{Overview of SurgicalMamba.} Frames are encoded by a partially frozen ConvNeXt and processed by $K$ stacked Dual-Path blocks, each running a Slow and a Fast SSD path plus a GELU FFN, before an Out Head produces per-frame phase predictions. \emph{Bottom left:} within a block, both paths apply a chunk-boundary rotation (\S\ref{sec:tstate}); the Slow path additionally carries state across clips, and an Intensity Net reads each frame to set $\lambda$, which lengthens the discretization step where a phase is ending so the state clears there rather than throughout (\S\ref{sec:slowpath}). The Slow path's output conditions the Fast path's selectivity. \emph{Bottom right:} the Out Head mirrors the Slow path without cross-clip carry.}
\label{fig:model}
\end{figure}
 
Online phase recognition produces, at each time $t$, a distribution
$\hat p_t \in \Delta^{C-1}$ over $C$ phases from frames $I_{1:t}$
only. For training, videos are split into non-overlapping clips of
length $L$, with slow-path state carried clip-to-clip under truncated
back-propagation through time (TBPTT); we write $\Delta_n =
\Delta(x_n)$ for the input-dependent step and subscript quantities
\emph{slow}/\emph{fast} by path.

SurgicalMamba keeps the standard encoder--temporal-model--classifier
recipe and redesigns the temporal model (Fig.~\ref{fig:model}). Two
demands pull the recurrence in opposite directions. Telling apart
phases that look alike needs context from minutes earlier, hence a
state that survives clip boundaries. Reacting to a brief event needs a
state not already committed to that history. The block
therefore runs two SSD paths on the same input with independent
projections and scans, one carrying its state across clips
(\S\ref{sec:slowpath}) and one resetting at every clip boundary
(\S\ref{sec:fastpath}), with the first conditioning the second. Within
each path, the discretization step of the slow one is warped by a
learned intensity (\S\ref{sec:slowpath}) and the carried state is
rotated at every chunk boundary (\S\ref{sec:tstate}). The surrounding
pipeline and all hyperparameters are standard and given in
\S\ref{sec:setup}.

\subsection{Slow path and intensity-modulated stepping}
\label{sec:slowpath}

Carrying state across clips makes the forward pass behave as if clips
were concatenated while gradients stay bounded by the TBPTT window
(Appendix~\ref{app:impl}). The second departure from a standard
Mamba2 block is that this path's discretization step is warped by a
learned intensity.

We treat $\lambda$ as a change of time variable, not an ad-hoc
multiplier on $\Delta$. Positing an intrinsic event time that runs
faster where the workflow is changing and at wall-clock rate where it
is not,
\begin{equation}
\tau(t) := \int_0^t \bigl(1+\lambda(s)\bigr)\,\mathrm ds,
\qquad
\alpha(t) := \frac{\mathrm d\tau}{\mathrm dt} = 1+\lambda(t) \in [1,2],
\label{eq:eventtime}
\end{equation}
and assuming the dynamics are stationary in $\tau$, the chain rule
re-expresses them in wall-clock time as $\mathrm dh/\mathrm
dt=\alpha(t)[A\,h(t)+B(t)x(t)]$. Since $\lambda$ is constant on each
frame interval ($\alpha_n:=1+\lambda_n$), exact ZOH integration gives
\begin{equation}
\bar A_n = \exp(\alpha_n\Delta A),
\qquad
\bar B_n = (\alpha_n\Delta A)^{-1}\bigl(\exp(\alpha_n\Delta A)-I\bigr)\,\alpha_n\Delta B_n
\;\approx\; \alpha_n\Delta B_n,
\label{eq:warpdisc}
\end{equation}
which is \emph{identical} to the standard Mamba2 discretization with
$\Delta$ replaced by $\alpha_n\Delta$: the SSD kernel, the structured
matrices, and the state shape are unchanged, and only the scalar step
is reweighted. We
instantiate
\begin{equation}
\lambda(t) = \sigma\bigl(\text{MLP}_\lambda(\text{sg}[\,x_\text{slow}^\text{conv}(t)\,])\bigr) \in [0,1]
\label{eq:lambda}
\end{equation}
with a small MLP (the \emph{Intensity Net} of
Fig.~\ref{fig:model}). Its only input is the depthwise-convolved frame
feature, so $\lambda$ has no access to elapsed time, to the phase
index, or to previous predictions: it must read a transition from
local visual evidence rather than from where in a procedure the model
happens to be. The input is detached ($\text{sg}[\cdot]$), so
$\mathcal{L}_\text{trans}$ trains the Intensity Net alone and
$\lambda$ reaches the rest of the model only through $\alpha_n\Delta$.
The ablation of \S\ref{sec:ablation} therefore removes the time-warp
itself, and not a representation that boundary supervision has shaped.

A single decay rate must both retain information within a phase and
forget it at a boundary. $\lambda$ separates these roles: it raises
the decay only near a boundary, so the base rate can be set solely by
how long information must be retained. A single frame at most doubles
its discretization step, so forgetting cannot be done in one frame and
must be spread over a short window.

Accordingly, $\lambda$ enters the recurrence through $\alpha_n\Delta$
alone, is predicted per frame from the convolved feature, and is
supervised toward a window centred on each phase boundary
(Fig.~\ref{fig:model}, \emph{Gaussian transition target}), wider after
the boundary than before: $\lambda$ is predicted causally, and
forgetting costs less on a segment that has just begun than on one
that has accumulated its full context. Appendix~\ref{app:lambda}
derives both. The warp is confined to the slow path,
whose cross-clip carry leaves a transient over-acceleration
recoverable.

\subsection{Fast path and fusion}
\label{sec:fastpath}

The fast path is a clip-local observer: its SSM state and conv buffer
reset at every clip boundary, so it reacts to within-clip events
without diluting its short-term selectivity or duplicating the slow
path's long-memory role. Within a clip it still propagates and rotates
state across chunk boundaries (\S\ref{sec:tstate}); only clip-level
carry is severed. Its selectivity is conditioned on the slow path: the fast
$(\Delta, B, C)$ are computed from the local frame together with the
slow path's long-context summary, so whether a grasper signals
dissection or retraction can depend on instruments seen minutes
earlier. Conditioning is one-way, letting the slow summary shape the
fast path's selectivity without writing transient events back into the
carried state. The fast scan is otherwise the standard Mamba2 form,
with no $\lambda$ modulation. The two outputs are summed channel-wise
and gated, which keeps the SSM inner dimension at the
canonical SSD value (Appendix~\ref{app:impl}).

\subsection{State regramming via Cayley rotation}
\label{sec:tstate}

The SSD scan advances the state
$h\in\mathbb{R}^{H\times P\times N}$ only through
$h\leftarrow \bar A\,h+\bar Bx$, with $\bar A$ a per-head scalar, so
the direction a frame occupies is fixed by $\bar B$ and therefore by
that frame alone: anything resembling it is written on
top of it. \emph{State regramming} rotates each head's state at chunk
boundaries by an input-conditioned orthogonal
$Z\in\mathbb{R}^{N\times N}$, so that the direction depends on the
chunks that have gone by as well. The same content arriving in a
different context is then held elsewhere and can be read out
separately, which Appendix~\ref{app:matrixview} works out in the SSD
matrix form.

At the end of chunk $c$ the per-head SSD output
$y^{(c)}\in\mathbb{R}^{B\times H\times P\times\Cchunk}$ is mean-pooled
over the chunk and LayerNorm-ed per head into a descriptor
\begin{equation}
\phi^{(c)} = \text{LN}_P\!\Bigl(\tfrac{1}{\Cchunk}\textstyle\sum_{t=1}^{\Cchunk} y^{(c)}[:,:,:,t]\Bigr)
\in \mathbb{R}^{B\times H\times P},
\label{eq:phi}
\end{equation}
which two per-head MLPs map to $r$ rotation planes and their angles,
\begin{equation}
[U \Vert V] = \text{MLP}_{UV}(\phi^{(c)}), \qquad
\theta = \text{softplus}\bigl(\text{MLP}_\theta(\phi^{(c)})\bigr) \in \mathbb{R}^{B\times H\times r}_{\ge0},
\label{eq:uvtheta}
\end{equation}
with $U,V\in\mathbb{R}^{B\times H\times N\times r}$ column-wise
$L_2$-normalized so that $\theta$ alone carries magnitude, letting the
chunk's content choose \emph{where} to rotate and \emph{by how much}
independently. The low-rank product is antisymmetrized,
\begin{equation}
\widetilde S = U\,\text{diag}(\theta)\,V^{\!\top}, \qquad
S^{(c)} = \widetilde S - \widetilde S^{\!\top},
\label{eq:skew}
\end{equation}
and the Cayley transform of the resulting skew-symmetric $S^{(c)}$
gives the rotation in closed form,
\begin{equation}
Z^{(c)} = \bigl(I - \tfrac{1}{2}S^{(c)}\bigr)^{-1}\bigl(I + \tfrac{1}{2}S^{(c)}\bigr) \in O(N),
\qquad
h^{(c)} \;\leftarrow\; h^{(c)}\,Z^{(c)}.
\label{eq:cayley}
\end{equation}
We use the Cayley form rather than a matrix exponential because it is
differentiable everywhere, $I-\tfrac12 S$ being invertible for any real
skew-symmetric $S$, and because it is orthogonal by construction
rather than approximately so, which matters when the slow path
composes these rotations over tens of chunk boundaries. Initializing
$\theta\approx0$ gives $Z^{(c)}\approx I$, so training starts from
vanilla SSD.

Rotating the state alone is what has an effect. The per-head scalar
$A$ commutes with any orthogonal $Q$, so the scan is
rotation-equivariant: turning $h$, $B$ and $C$ together leaves the
output unchanged. Turning only $h$ therefore leaves the next chunk's
fresh $B^{(c+1)},C^{(c+1)}$ reading the carried state in a re-oriented
basis, and each chunk summarizes itself and chooses the basis in which
that summary is passed on. Nothing is lost in the process:
$\lVert h^{(c)}Z^{(c)}\rVert_F=\lVert h^{(c)}\rVert_F$, so the content
of the carry is redirected but not attenuated, and by the same
commutation the boundary rotation preserves the
$N$-semiseparable structure of SSD and its $O(d)$ per-frame inference
cost, which
Appendix~\ref{app:matrixview} states as
Proposition~\ref{prop:preserve}. Per-head batching and the cost of the
rotation, which is amortized over the frames of a chunk, are in
Appendix~\ref{app:impl}.

Conditioning $Z$ on content is a requirement of the mechanism rather
than one of its settings. Were $Z$ the same at every boundary, the
product accumulated between chunks $c'$ and $c$ would reduce to
$Z^{\,c-c'}$, which factors as $Z^{-c'}\!\cdot Z^{c}$; the same
equivariance then absorbs the two factors into the chunk-$c'$ writes
and the chunk-$c$ reads. What is left is vanilla SSD with a
chunk-level relative position encoding on $B$ and $C$, and the kernel
factorizes once more through the two frames and the chunk distance
between them, which is the form
Appendix~\ref{app:matrixview} identifies as the limitation. Only a
$Z$ that varies with content makes the accumulated product depend on
what lay between two frames rather than on how far apart they are.

\subsection{Output head and training objectives}
\label{sec:loss}

A lightweight SSD-augmented head maps the last block's output to
per-frame logits (Fig.~\ref{fig:model}, bottom right), mirroring the
slow path without cross-clip carry: state resets each clip, rotations
are applied within it, and the gated result is projected to $C$ phase
classes. The clip loss is
$\mathcal{L}=\mathcal{L}_\text{CE}+w_\text{sm}\mathcal{L}_\text{smooth}
+w_\text{trans}\mathcal{L}_\text{trans}$, over valid (non-padded)
positions.

$\mathcal{L}_\text{CE}$ is cross-entropy with label smoothing.
$\mathcal{L}_\text{smooth}$ penalizes frame-to-frame change in the
prediction, but weighted by the confidence at both endpoints
($c_{b,t}=1-H(\hat p_{b,t})/\log C$), so that it acts within a phase,
where the model is confident, and relaxes at transitions, where
predictions \emph{should} change abruptly. $\mathcal{L}_\text{trans}$
is a per-layer BCE between each layer's intensity $\lambda^{(\ell)}$
and the transition target of \S\ref{sec:slowpath}, an asymmetric
Gaussian peaking at $1$ on each boundary with standard deviations
$\sigma_L<\sigma_R$ before and after (values in
Appendix~\ref{app:impl}); targets of this kind are also used for
boundary supervision by LoViT~\cite{liu2025lovit}.

\section{Experiments}
\label{sec:experiments}
 
\subsection{Setup}
\label{sec:setup}

We evaluate on seven public surgical-video datasets. The main text
reports \textbf{Cholec80}~\cite{twinanda2016endonet} and
\textbf{M2CAI16}~\cite{twinanda2016m2cai}, both laparoscopic
cholecystectomy, and \textbf{AutoLaparo}~\cite{wang2022autolaparo},
laparoscopic hysterectomy; the remaining four span cataract,
multi-procedure, and prostatectomy surgery. All are sub-sampled to
$1$\,fps~\cite{twinanda2016endonet, jin2018svrcnet} and evaluated
causally: at each frame the model predicts from $I_{1:t}$ only, with
per-frame work independent of look-back length. We report video-wise
accuracy and phase-wise precision, recall and
Jaccard~\cite{jin2018svrcnet, funke2023metrics, huang2025mttrnet},
emphasizing Jaccard as the metric most sensitive to hard-to-recognize
phases, and for Cholec80 report both with and without the 10-second
relaxed boundary~\cite{twinanda2016m2cai}. Dataset splits and metric
definitions are in Appendix~\ref{app:morebench}; the backbone,
hyperparameters, and their sensitivity are in
Appendices~\ref{app:impl} and~\ref{app:hparam}.

\subsection{Comparison with state-of-the-art methods}
\label{sec:sota}
 
\begin{table}[t]
\centering
\caption{Comparison with state-of-the-art methods on the three
main-text benchmarks, each under its standard protocol. Best per block
in bold; ``--'' not reported by the source.}
\label{tab:main}
\small
\renewcommand{\arraystretch}{0.9}
\setlength{\tabcolsep}{7pt}
\begin{tabular}{lcccc}
\toprule
Method & Accuracy & Precision & Recall & Jaccard \\
\midrule
\multicolumn{5}{l}{\textit{Relaxed (10-second boundary)}} \\
\midrule
PhaseNet~\cite{twinanda2016endonet}      & 78.8\pms{4.7} & 71.3\pms{15.6} & 76.6\pms{16.6} & -- \\
EndoNet~\cite{twinanda2016endonet}       & 81.7\pms{4.2} & 73.7\pms{16.1} & 79.6\pms{7.9} & -- \\
SV-RCNet~\cite{jin2018svrcnet}           & 85.3\pms{7.3} & 80.7\pms{7.0} & 83.5\pms{7.5} & -- \\
MTRCNet-CL~\cite{jin2020mtrcnet}         & 89.2\pms{7.6} & 86.9\pms{4.3} & 88.0\pms{6.9} & -- \\
TeCNO~\cite{czempiel2020tecno}           & 88.6\pms{7.8} & 81.6\pms{7.0} & 85.2\pms{6.7} & 75.1\pms{6.9} \\
Opera~\cite{czempiel2021opera}           & 91.2\pms{6.4} & 82.2\pms{7.0} & 86.9\pms{8.6} & -- \\
TMRNet~\cite{jin2021tmrnet}              & 90.1\pms{7.6} & 90.3\pms{3.3} & 89.5\pms{5.0} & 79.1\pms{5.7} \\
Trans-SVNet~\cite{gao2021transsvnet}     & 90.3\pms{7.1} & 90.7\pms{5.0} & 88.8\pms{7.4} & 79.3\pms{6.6} \\
Not E2E~\cite{yi2022note2e}              & 91.5\pms{7.1} & -- & 86.8\pms{8.5} & 77.2\pms{11.2} \\
UATD~\cite{ding2023uatd}                 & 91.9\pms{5.6} & 89.5\pms{4.4} & 90.5\pms{5.9} & 79.9\pms{8.5} \\
CMTNet~\cite{yue2023cmtnet}              & 92.9\pms{5.9} & 90.1\pms{7.1} & 92.0\pms{4.4} & 81.5\pms{10.4} \\
LoViT~\cite{liu2025lovit}                & 92.4\pms{6.3} & 89.9\pms{6.1} & 90.6\pms{4.4} & 81.2\pms{9.1} \\
SKiT~\cite{liu2023skit}                  & 93.4\pms{5.2} & 90.9 & 91.8 & 82.6 \\
SR-Mamba~\cite{cao2024srmamba}           & 92.6\pms{8.6} & 90.3\pms{5.2} & 90.6\pms{7.2} & 81.5\pms{8.6} \\
Surgformer~\cite{yang2024surgformer}     & 93.4\pms{6.4} & 91.9\pms{4.7} & 92.1\pms{5.8} & 84.1\pms{8.0} \\
DACAT~\cite{yang2024dacat}               & 95.5\pms{4.3} & 93.6\pms{4.1} & 93.4\pms{5.3} & 87.4\pms{8.1} \\
MTTR-Net~\cite{huang2025mttrnet}         & 95.0\pms{4.9} & 92.7\pms{5.1} & 91.2\pms{10.3} & 84.7\pms{11.0} \\
\textbf{SurgicalMamba (Ours)}            & \textbf{96.1}\pms{3.6} & \textbf{94.9}\pms{4.2} & \textbf{94.4}\pms{6.2} & \textbf{88.5}\pms{8.1} \\
\midrule
\multicolumn{5}{l}{\textit{Strict (unrelaxed)}} \\
\midrule
Trans-SVNet~\cite{gao2021transsvnet}     & 89.1\pms{7.0} & 84.7 & 83.6 & 72.5 \\
LoViT~\cite{liu2025lovit}                & 91.5\pms{6.1} & 83.1 & 86.5 & 74.2 \\
SKiT~\cite{liu2023skit}                  & 92.5\pms{5.1} & 84.6 & 88.5 & 76.7 \\
Surgformer~\cite{yang2024surgformer}     & 92.4\pms{6.4} & 87.9\pms{6.9} & 89.3\pms{7.8} & 79.9\pms{10.2} \\
MTTR-Net~\cite{huang2025mttrnet}         & 93.9\pms{5.0} & 88.8\pms{6.9} & 88.2\pms{10.8} & 80.5\pms{11.8} \\
\textbf{SurgicalMamba (Ours)}            & \textbf{94.6}\pms{3.7} & \textbf{89.6}\pms{8.7} & \textbf{90.5}\pms{8.1} & \textbf{82.7}\pms{11.5} \\
\midrule
\multicolumn{5}{l}{\textit{M2CAI16} (relaxed, 10-second boundary)} \\
\midrule
TMRNet~\cite{jin2021tmrnet}              & 87.0\pms{8.6} & 87.8\pms{6.9} & 88.4\pms{5.3} & 75.1\pms{6.9} \\
Trans-SVNet~\cite{gao2021transsvnet}     & 87.2\pms{9.3} & 88.0\pms{6.7} & 87.5\pms{5.5} & 74.7\pms{7.7} \\
UATD~\cite{ding2023uatd}                 & 87.6\pms{8.7} & 88.2\pms{7.4} & 87.9\pms{9.6} & 75.7\pms{9.5} \\
CMTNet~\cite{yue2023cmtnet}              & 88.2\pms{9.2} & 88.3\pms{7.8} & 88.7\pms{6.2} & 76.1\pms{9.2} \\
DACAT~\cite{yang2024dacat}               & 91.3\pms{9.3} & 90.8\pms{7.6} & 90.6\pms{6.7} & 80.7\pms{8.8} \\
\textbf{SurgicalMamba (Ours)}            & \textbf{92.2}\pms{8.8} & \textbf{91.8}\pms{7.2} & \textbf{91.4}\pms{7.7} & \textbf{83.3}\pms{9.7} \\
\midrule
\multicolumn{5}{l}{\textit{AutoLaparo} (strict)} \\
\midrule
LoViT~\cite{liu2025lovit}                & 81.4\pms{7.6} & 85.1 & 65.9 & 55.9 \\
SKiT~\cite{liu2023skit}                  & 82.9\pms{6.8} & 81.8 & 70.1 & 59.9 \\
Surgformer~\cite{yang2024surgformer}     & 85.7\pms{6.9} & 82.3 & 75.7 & 66.7 \\
MTTR-Net~\cite{huang2025mttrnet}         & 85.4\pms{9.2} & 78.8 & 76.8 & 65.6 \\
DACAT~\cite{yang2024dacat}               & 87.8\pms{7.6} & 78.5 & 75.0 & 66.9 \\
\textbf{SurgicalMamba (Ours)}            & \textbf{89.5}\pms{6.8} & \textbf{90.6} & \textbf{76.2} & \textbf{68.9} \\
\bottomrule
\end{tabular}
\end{table}

We compare against results reported in the original papers, all under
the protocol of \S\ref{sec:setup} (Table~\ref{tab:main}). On
Cholec80 SurgicalMamba is state-of-the-art on every metric under both
protocols, with the gain concentrated on Jaccard, which improves by
$+1.1$\,pp relaxed and $+2.2$\,pp strict over the strongest prior
method. The same pattern holds on the other two main-text benchmarks,
with $+2.6$\,pp Jaccard over DACAT on M2CAI16, and on AutoLaparo a large
precision gain ($90.6$ against $78.5\%$) alongside the best accuracy
and Jaccard. The four remaining benchmarks
(Appendix~\ref{app:morebench}) follow suit, with the widest margins on
the smallest and most domain-shifted sets, and our per-video standard
deviations are among the lowest reported.

\begin{table}[t]
\centering
\caption{Component ablation on Cholec80 (\%, mean$\pm$std over 40
videos), under both protocols. Shaded rows are controls:
\emph{Vanilla Mamba2} is a single-path baseline and \emph{+ $Z$ only}
adds state regramming alone; the middle rows remove one component from
the \emph{Full} model. Best in bold.}
\label{tab:component_ablation}
\scriptsize
\renewcommand{\arraystretch}{0.9}
\setlength{\tabcolsep}{3.5pt}
\begin{tabular}{lcccccccc}
\toprule
& \multicolumn{4}{c}{Relaxed (10-s boundary)} & \multicolumn{4}{c}{Strict} \\
\cmidrule(lr){2-5}\cmidrule(lr){6-9}
Variant & Accuracy & Precision & Recall & Jaccard & Accuracy & Precision & Recall & Jaccard \\
\midrule
\rowcolor{gray!10} Vanilla Mamba2 & 93.82\pms{5.09} & 92.02\pms{6.30} & 90.47\pms{10.35} & 83.37\pms{10.62} & 92.45\pms{5.23} & 86.97\pms{11.50} & 86.16\pms{12.38} & 77.69\pms{13.57} \\
\rowcolor{gray!10} + $Z$ only & 94.70\pms{4.29} & 92.46\pms{5.10} & 93.05\pms{6.11} & 85.08\pms{9.25} & 93.16\pms{4.48} & 86.97\pms{9.78} & 88.84\pms{7.44} & 79.21\pms{12.11} \\
w/o rotation $Z$ & 95.60\pms{4.20} & 92.97\pms{4.25} & 93.71\pms{7.95} & 87.05\pms{9.05} & 94.01\pms{4.28} & 88.07\pms{7.68} & 89.77\pms{9.82} & 80.71\pms{12.07} \\
w/o intensity $\lambda$ & 95.08\pms{6.07} & 93.37\pms{4.47} & 93.49\pms{5.04} & 86.70\pms{8.75} & 93.69\pms{6.05} & 88.72\pms{8.06} & 89.26\pms{7.30} & 81.12\pms{11.99} \\
w/o fast path & 95.85\pms{4.03} & 94.28\pms{4.03} & 93.47\pms{8.65} & 88.03\pms{9.44} & 94.27\pms{4.14} & 88.92\pms{8.91} & 88.67\pms{10.84} & 81.24\pms{12.69} \\
w/o smooth loss & 95.84\pms{3.69} & 93.54\pms{4.71} & 94.43\pms{5.41} & 87.70\pms{8.48} & 94.38\pms{3.82} & 88.55\pms{8.59} & 90.36\pms{7.80} & 81.80\pms{11.81} \\
\textbf{Full} & \textbf{96.05}\pms{3.55} & \textbf{94.91}\pms{4.24} & \textbf{94.38}\pms{6.16} & \textbf{88.48}\pms{8.14} & \textbf{94.61}\pms{3.71} & \textbf{89.59}\pms{8.70} & \textbf{90.48}\pms{8.11} & \textbf{82.73}\pms{11.50} \\
\bottomrule
\end{tabular}
\end{table}

\subsection{Ablation Studies}
\label{sec:ablation}
 
On Cholec80 we remove one component at a time from the full model, and
separately build up from a vanilla Mamba2 baseline
(Table~\ref{tab:component_ablation}). The two controls bracket the
joint effect: a vanilla single-path Mamba2 reaches $77.7\%$ strict
Jaccard against the full model's $82.7\%$, and state regramming alone
recovers $+1.5$\,pp of that gap.

Each component contributes under both protocols, and the three main
ones do so on the same order. Removing $Z$ costs $2.0$\,pp strict
Jaccard, the intensity $1.6$, and the fast path $1.5$, with the
smoothness loss adding $0.9$. These differences are smaller than the
inter-video spread. What separates $\lambda$ is not the size of its
drop but its effect on consistency: removing it inflates the per-video
accuracy standard deviation from $3.71$ to $6.05$, where removing $Z$
or the fast path leaves it near $4.2$. That is what transition-aware
supervision is for, and not something a generic loss of capacity would
produce.

\subsection{Does the mechanism transfer beyond surgery?}
\label{sec:mqar}

\begin{wrapfigure}{r}{0.5\textwidth}
\vspace{-8pt}
\centering
\includegraphics[width=0.48\textwidth]{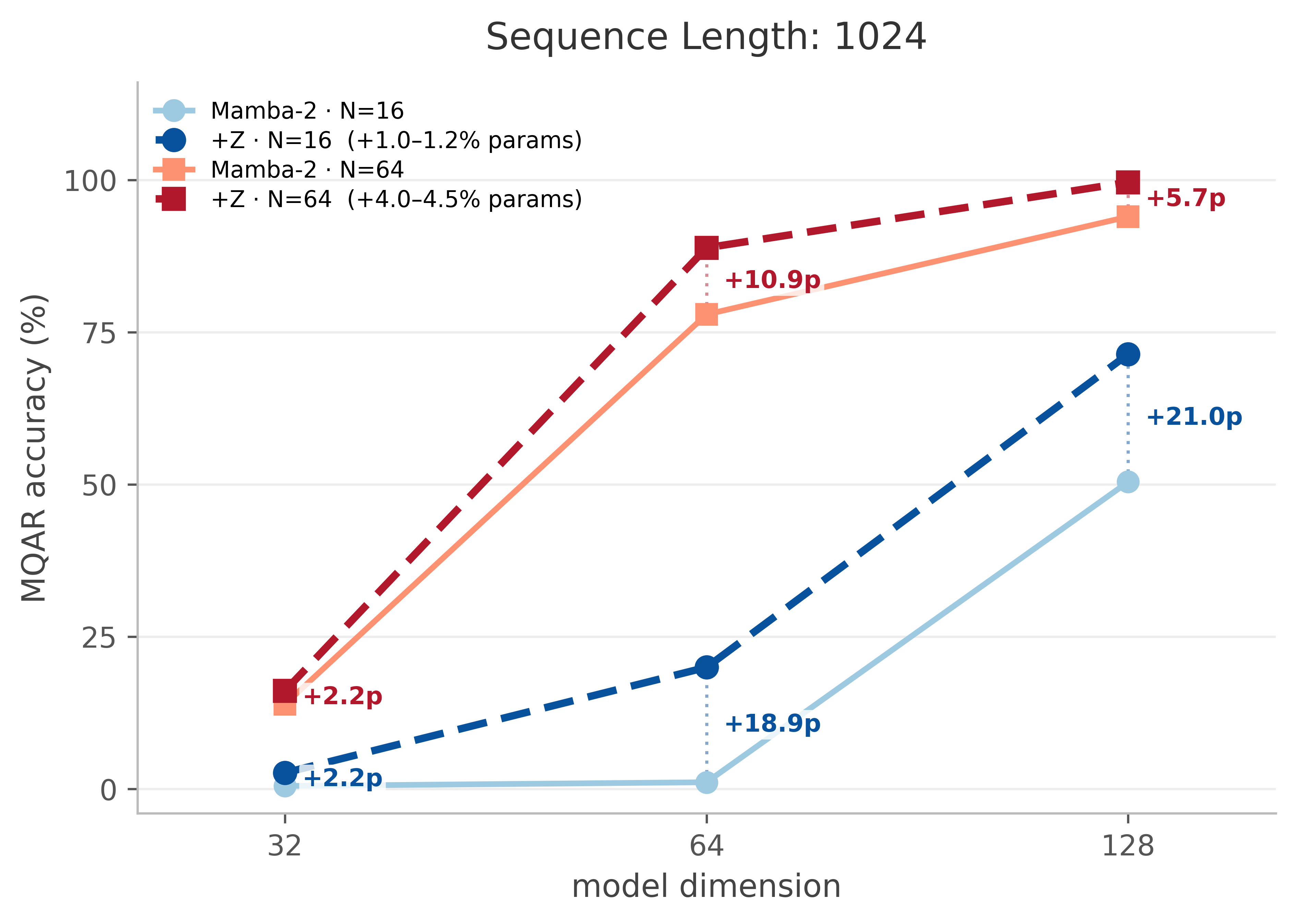}
\caption{MQAR at sequence length $1024$ ($256$ key--value pairs).
Adding $Z$ improves every configuration that is not already saturated,
and most where the state is scarcest relative to what must be stored.
Sequence lengths $256$ and $512$, and the protocol, are in
Appendix~\ref{app:mqar}.}
\label{fig:mqar}
\end{wrapfigure}

Multi-query associative recall (MQAR) tests the claim of
\S\ref{sec:tstate} where surgical structure cannot be responsible for
the outcome: key--value pairs are scattered through a sequence of
distractors and must later be retrieved one at a time, so a run
succeeds or fails on whether the state kept them apart. The context
separating two pairs is then only their position in the sequence,
which is what makes the test a clean one. It is also the diagnostic
Mamba2 used to argue for a larger recurrent
state~\cite{dao2024transformers}, the alternative remedy for the same
problem: if similar keys are written on top of one another, one answer
is to keep enlarging the state until they stop colliding. We reuse
that protocol unmodified (Appendix~\ref{app:mqar}) and change only the
boundary rotation; $\lambda$ is excluded, as the task supplies no
transitions to supervise it.

The rotation improves every configuration that is not already
saturated, for $1$--$4.5\%$ more parameters, and the gain is largest
where the state is scarcest relative to the number of pairs. Capacity
added to the model would have helped uniformly, including where the
state was never the constraint; something that only decides
\emph{where} content is placed can help only while placement is
contested. Enlarging the state remains the stronger remedy on this
task, so the rotation complements rather than replaces it.

\subsection{Streaming efficiency}
\label{sec:efficiency}

\begin{table}[t]
\centering
\caption{Model forward at batch $1$ on one RTX A6000, averaged over
$500$ frames.}
\label{tab:efficiency}
\small
\setlength{\tabcolsep}{10pt}
\renewcommand{\arraystretch}{1.2}
\begin{tabular*}{\textwidth}{@{\extracolsep{\fill}}ccccrr@{}}
\toprule
\multicolumn{4}{c}{Model} & \multicolumn{2}{c}{Throughput} \\
\cmidrule(r){1-4}\cmidrule(l){5-6}
Parameters & GFLOPs / frame & Peak memory & Per-frame cost
& Standard & Graph capture \\
\midrule
198.56\,M & 4.55 & $<1$\,GB & $O(d)$ & 88.42\,fps & \textbf{312.88\,fps} \\
\bottomrule
\end{tabular*}
\end{table}

Measuring the model forward at batch $1$, SurgicalMamba reaches
$312.88$\,fps with CUDA graph capture and $88.42$\,fps without it
(Table~\ref{tab:efficiency}), both above typical $25$--$60$\,fps
endoscopic capture rates. What matters for streaming is not the rate
itself but its behavior in $T$: per-frame work is $O(d)$ and the
carried state has fixed size, so these are steady-state rates that
hold however long an operation runs. The same property is what makes
graph capture applicable at all, since the computation issued per
frame does not change from one frame to the next and can be captured
once and replayed.

\subsection{Qualitative analysis}
\label{sec:qualitative}

\begin{figure}[t]
\centering
\includegraphics[width=0.9\textwidth]{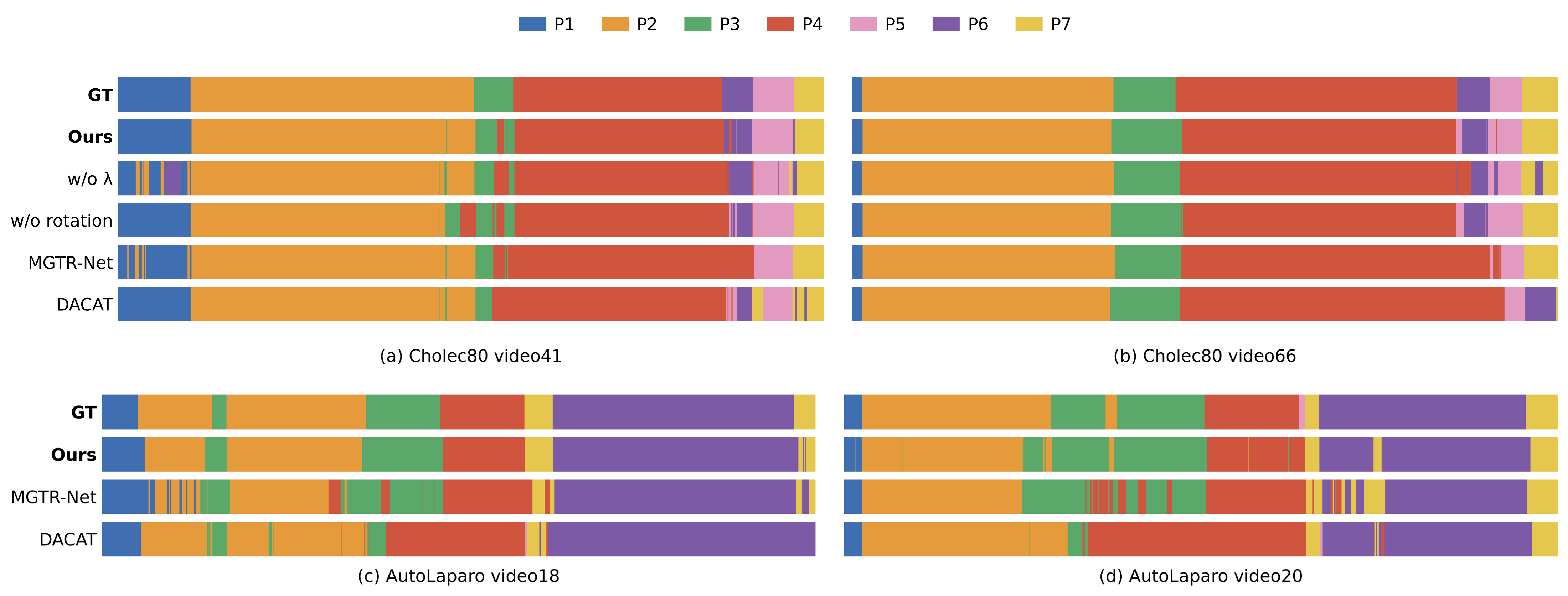}
\caption{Predicted phase timelines on four test videos (frame index on
the horizontal axis; phases colour-coded). Ablation rows are shown for
Cholec80, where both mechanisms are trained. Internal views are in
Appendix~\ref{app:qual}.}
\label{fig:qual}
\end{figure}

Figure~\ref{fig:qual} shows predicted phase timelines for four test
videos across two datasets, comparing SurgicalMamba against the ground
truth, the no-$\lambda$ and no-rotation ablations, and the strongest
baselines. Both baselines over-segment early:
MTTR-Net~\cite{huang2025mttrnet} fragments the opening \emph{Prep}
phase in (a) and scatters throughout (c) and (d), and DACAT flickers
around the late short phases. Removing $\lambda$ reproduces exactly
that early-onset instability in (a), the transition it is supervised
to sharpen. Removing $Z$ leaves onsets intact but fragments the
\emph{ClipCut}/\emph{GBDiss} transition in (a), where two visually
similar phases meet. This is the per-video form of the strict-Jaccard
gap in Table~\ref{tab:main}. The pattern is consistent across
procedures. AutoLaparo in (c) and (d) shows the same baseline
fragmentation in the
long \emph{DivLigament} and \emph{DivUterine} stretches, while our
predictions stay contiguous.
Appendix~\ref{app:qual} gives two internal views of the same video,
where the rotation planes stay aligned within a phase and turn at its
end, and the carried state occupies a visibly wider portion of its
space than without $Z$.

\section{Conclusion}
\label{sec:conclusion}

The SSD form owes its efficiency to the scan seeing a per-head scalar
transition, so the state it carries is only rescaled from step to
step, never re-arranged. The step that sets the rate does vary with
the input, but prior work reports that this variation is not reliably
learned into a forgetting schedule~\cite{chen2024stuffed,
wang2025memmamba}. Neither property is specific to one model, and both
sit badly with surgical video. The same instruments and views recur
throughout a procedure, so repeated content is written into the same
directions. Procedures are long and change little visually from frame
to frame, and their phases differ widely in length, so there is little
for the step to select on and no fixed rate that would serve instead.
This work addresses both at the state handed across chunk boundaries,
rather than at the transition itself, by applying an
input-conditioned rotation together with a supervised warping of the
discretization step. Both operations preserve $N$-semiseparability and
$O(d)$ per-frame cost. Across seven online SPR benchmarks the model
achieves state-of-the-art accuracy and Jaccard at $312.88$\,fps, and
it also improves multi-query associative recall in a setting without
surgical structure.

These results indicate that a fixed-cost recurrence can be adapted to
a domain through its state hand-off, without sacrificing the cost
profile that motivated its use. Placing $\lambda$ does require knowing
where the phase boundaries are; in tasks without such annotations they
would have to be inferred, or another criterion found for when the
state should be cleared.

\bibliographystyle{plainnat}
\bibliography{references}

@article{maierhein2017sds,
  title={Surgical data science for next-generation interventions},
  author={Maier-Hein, Lena and Vedula, Swaroop S and Speidel, Stefanie and Navab, Nassir and Kikinis, Ron and Park, Adrian and Eisenmann, Matthias and Feussner, Hubertus and Forestier, Germain and Giannarou, Stamatia and others},
  journal={Nature Biomedical Engineering},
  volume={1}, number={9}, pages={691--696}, year={2017}
}

@article{garrow2021ml,
  title={Machine learning for surgical phase recognition: a systematic review},
  author={Garrow, Carly R and Kowalewski, Karl-Friedrich and Li, Linhong and Wagner, Martin and Schmidt, Mona W and Engelhardt, Sandy and Hashimoto, Daniel A and Kenngott, Hannes G and Bodenstedt, Sebastian and Speidel, Stefanie and others},
  journal={Annals of Surgery},
  volume={273}, number={4}, pages={684--693}, year={2021}
}

@article{twinanda2016endonet,
  title={EndoNet: a deep architecture for recognition tasks on laparoscopic videos},
  author={Twinanda, Andru P and Shehata, Sherif and Mutter, Didier and Marescaux, Jacques and De Mathelin, Michel and Padoy, Nicolas},
  journal={IEEE Transactions on Medical Imaging},
  volume={36}, number={1}, pages={86--97}, year={2016}
}

@article{wagner2023heichole,
  title={Comparative validation of machine learning algorithms for surgical workflow and skill analysis with the {HeiChole} benchmark},
  author={Wagner, Martin and M{\"u}ller-Stich, Beat-Peter and Kisilenko, Anna and Tran, Duc and Heger, Patrick and M{\"u}ndermann, Lars and Lubotsky, David M and Mu{\"u}ller, Benjamin and Davitashvili, Tornike and Capek, Manuela and others},
  journal={Medical Image Analysis},
  volume={86}, pages={102770}, year={2023}
}

@article{maierhein2021heico,
  title={Heidelberg colorectal data set for surgical data science in the sensor operating room},
  author={Maier-Hein, Lena and Wagner, Martin and Ross, Tobias and Reinke, Annika and Bodenstedt, Sebastian and Full, Peter M and Hempe, Hellena and Mindroc-Filimon, Diana and Scholz, Patrick and Tran, Thuy Nuong and others},
  journal={Scientific Data},
  volume={8}, number={1}, pages={101}, year={2021}
}

@article{jin2018svrcnet,
  title={{SV-RCNet}: workflow recognition from surgical videos using recurrent convolutional network},
  author={Jin, Yueming and Dou, Qi and Chen, Hao and Yu, Lequan and Qin, Jing and Fu, Chi-Wing and Heng, Pheng-Ann},
  journal={IEEE Transactions on Medical Imaging},
  volume={37}, number={5}, pages={1114--1126}, year={2018}
}

@article{jin2020mtrcnet,
  title={Multi-task recurrent convolutional network with correlation loss for surgical video analysis},
  author={Jin, Yueming and Li, Huaxia and Dou, Qi and Chen, Hao and Qin, Jing and Fu, Chi-Wing and Heng, Pheng-Ann},
  journal={Medical Image Analysis},
  volume={59}, pages={101572}, year={2020}
}

@inproceedings{czempiel2020tecno,
  title={{TeCNO}: Surgical phase recognition with multi-stage temporal convolutional networks},
  author={Czempiel, Tobias and Paschali, Magdalini and Keicher, Matthias and Simson, Walter and Feussner, Hubertus and Kim, Seong Tae and Navab, Nassir},
  booktitle={MICCAI}, pages={343--352}, year={2020}
}

@inproceedings{czempiel2021opera,
  title={{OperA}: Attention-regularized transformers for surgical phase recognition},
  author={Czempiel, Tobias and Paschali, Magdalini and Ostler, Daniel and Kim, Seong Tae and Busam, Benjamin and Navab, Nassir},
  booktitle={MICCAI}, pages={604--614}, year={2021}
}

@article{rivoir2024bnpitfalls,
  title={On the pitfalls of batch normalization for end-to-end video learning: a study on surgical workflow analysis},
  author={Rivoir, Dominik and Funke, Isabel and Speidel, Stefanie},
  journal={Medical Image Analysis},
  volume={94}, pages={103126}, year={2024}
}

@article{jin2021tmrnet,
  title={Temporal memory relation network for workflow recognition from surgical video},
  author={Jin, Yueming and Long, Yonghao and Chen, Cheng and Zhao, Zixu and Dou, Qi and Heng, Pheng-Ann},
  journal={IEEE Transactions on Medical Imaging},
  volume={40}, number={7}, pages={1911--1923}, year={2021}
}

@inproceedings{yang2024dacat,
  title={{DACAT}: Dual-stream adaptive clip-aware time modeling for robust online surgical phase recognition},
  author={Yang, Kaixiang and Li, Qiang and Wang, Zhiwei},
  booktitle={ICASSP}, year={2025}
}

@inproceedings{gao2021transsvnet,
  title={Trans-SVNet: accurate phase recognition from surgical videos via hybrid embedding aggregation transformer},
  author={Gao, Xiaojie and Jin, Yueming and Long, Yonghao and Dou, Qi and Heng, Pheng-Ann},
  booktitle={MICCAI}, year={2021}
}

@article{yue2023cmtnet,
  title={Cascade multi-level transformer network for surgical workflow analysis},
  author={Yue, Wenkang and Liao, Hu and Xia, Yuan and Lam, Vivian and Luo, Jiebo and Wang, Zhiwei},
  journal={IEEE Transactions on Medical Imaging},
  volume={42}, number={10}, pages={2817--2831}, year={2023}
}

@inproceedings{liu2023skit,
  title={{SKiT}: a fast key information video transformer for online surgical phase recognition},
  author={Liu, Yang and Huo, Jiayu and Peng, Jingjing and Sparks, Rachel and Dasgupta, Prokar and Granados, Alejandro and Ourselin, Sebastien},
  booktitle={ICCV}, pages={21074--21084}, year={2023}
}

@inproceedings{yang2024surgformer,
  title={Surgformer: Surgical transformer with hierarchical temporal attention for surgical phase recognition},
  author={Yang, Shu and Luo, Luyang and Wang, Qi and Chen, Hao},
  booktitle={MICCAI}, pages={606--616}, year={2024}
}

@article{liu2025lovit,
  title={{LoViT}: Long video transformer for surgical phase recognition},
  author={Liu, Yang and Boels, Maxence and Garcia-Peraza-Herrera, Luis C and Vercauteren, Tom and Dasgupta, Prokar and Granados, Alejandro and Ourselin, Sebastien},
  journal={Medical Image Analysis}, year={2025}
}

@article{huang2025mttrnet,
  title={Multi-Teacher Temporal Regulation Network for Surgical Workflow Recognition},
  author={Huang, Kaide and Yuan, Xiang-Lei and Liu, Rui-De and Ye, Lian-Song and Zhou, Yao and Hu, Bing and Yi, Zhang},
  journal={IEEE Transactions on Medical Imaging},
  volume={44}, number={11}, pages={4690--4703}, year={2025}
}

@inproceedings{gu2022s4,
  title={Efficiently modeling long sequences with structured state spaces},
  author={Gu, Albert and Goel, Karan and R{\'e}, Christopher},
  booktitle={ICLR}, year={2022}
}

@inproceedings{gu2023mamba,
  title={Mamba: Linear-time sequence modeling with selective state spaces},
  author={Gu, Albert and Dao, Tri},
  booktitle={Conference on Language Modeling (COLM)}, year={2024}
}

@inproceedings{dao2024transformers,
  title={Transformers are {SSMs}: Generalized models and efficient algorithms through structured state space duality},
  author={Dao, Tri and Gu, Albert},
  booktitle={ICML}, year={2024}
}

@article{cao2024srmamba,
  title={{SR-Mamba}: Effective surgical phase recognition with state space model},
  author={Cao, Rui and Wang, Jian and Liu, Yun-Hui},
  journal={arXiv preprint arXiv:2407.08333}, year={2024}
}

@article{chen2024stuffed,
  title={Stuffed mamba: Oversized states lead to the inability to forget},
  author={Chen, Yingfa and Zhang, Xinrong and Hu, Shengding and Han, Xu and Liu, Zhiyuan and Sun, Maosong},
  journal={arXiv preprint arXiv:2410.07145}, year={2024}
}

@article{wang2025memmamba,
  title={{MemMamba}: Rethinking memory patterns in state space model},
  author={Wang, Youjin and Chen, Yangjingyi and Yan, Jiahao and Lu, Jiaxuan and Sun, Xiao},
  journal={arXiv preprint arXiv:2510.03279}, year={2025}
}

@inproceedings{arjovsky2016urnn,
  title={Unitary evolution recurrent neural networks},
  author={Arjovsky, Martin and Shah, Amar and Bengio, Yoshua},
  booktitle={ICML}, year={2016}
}

@inproceedings{mhammedi2017ornn,
  title={Efficient orthogonal parametrisation of recurrent neural networks using {Householder} reflections},
  author={Mhammedi, Zakaria and Hellicar, Andrew and Rahman, Ashfaqur and Bailey, James},
  booktitle={ICML}, year={2017}
}

@inproceedings{jing2017eunn,
  title={Tunable efficient unitary neural networks ({EUNN}) and their application to {RNNs}},
  author={Jing, Li and Shen, Yichen and Dub{\v{c}}ek, Tena and Peurifoy, John and Skirlo, Scott and LeCun, Yann and Tegmark, Max and Solja{\v{c}}i{\'c}, Marin},
  booktitle={ICML}, year={2017}
}

@inproceedings{liu2022convnext,
  title={A {ConvNet} for the 2020s},
  author={Liu, Zhuang and Mao, Hanzi and Wu, Chao-Yuan and Feichtenhofer, Christoph and Darrell, Trevor and Xie, Saining},
  booktitle={CVPR}, year={2022}
}

@inproceedings{feichtenhofer2019slowfast,
  title={{SlowFast} networks for video recognition},
  author={Feichtenhofer, Christoph and Fan, Haoqi and Malik, Jitendra and He, Kaiming},
  booktitle={ICCV}, year={2019}
}

@article{hashimoto2018ai,
  title={Artificial intelligence in surgery: promises and perils},
  author={Hashimoto, Daniel A and Rosman, Guy and Rus, Daniela and Meireles, Ozanan R},
  journal={Annals of Surgery},
  volume={268}, number={1}, pages={70--76}, year={2018}
}

@article{mascagni2022computer,
  title={Computer vision in surgery: from potential to clinical value},
  author={Mascagni, Pietro and Alapatt, Deepak and Sestini, Luca and Altieri, Maria S and Madani, Adnan and Watanabe, Yusuke and Alseidi, Adnoor and Redan, Jay A and Alfieri, Sergio and Costamagna, Guido and others},
  journal={npj Digital Medicine},
  volume={5}, number={1}, pages={163}, year={2022}
}

@inproceedings{helfrich2018orthogonal,
  title={Orthogonal recurrent neural networks with scaled {Cayley} transform},
  author={Helfrich, Kyle and Willmott, Devin and Ye, Qiang},
  booktitle={ICML}, year={2018}
}

@inproceedings{lezcano2019cheap,
  title={Cheap orthogonal constraints in neural networks: A simple parametrization of the orthogonal and unitary group},
  author={Lezcano-Casado, Mario and Mart{\'\i}nez-Rubio, David},
  booktitle={ICML}, year={2019}
}

@inproceedings{wang2022autolaparo,
  title={{AutoLaparo}: A new dataset of integrated multi-tasks for image-guided surgical automation in laparoscopic hysterectomy},
  author={Wang, Ziyi and Lu, Bo and Long, Yonghao and Zhong, Fangxun and Cheung, Tak-Hong and Dou, Qi and Liu, Yunhui},
  booktitle={MICCAI},
  pages={486--496},
  year={2022}
}

@misc{twinanda2016m2cai,
  title={Workshop and challenges on modeling and monitoring of computer assisted interventions ({M2CAI})},
  author={Twinanda, Andru P and Shehata, Sherif and Mutter, Didier and Marescaux, Jacques and De Mathelin, Michel and Padoy, Nicolas},
  year={2016},
  howpublished={\url{http://camma.u-strasbg.fr/m2cai2016/}}
}

@article{funke2023metrics,
  title={Metrics matter in surgical phase recognition},
  author={Funke, Isabel and Rivoir, Dominik and Speidel, Stefanie},
  journal={arXiv preprint arXiv:2305.13961},
  year={2023}
}

@inproceedings{he2016resnet,
  title={Deep residual learning for image recognition},
  author={He, Kaiming and Zhang, Xiangyu and Ren, Shaoqing and Sun, Jian},
  booktitle={CVPR}, pages={770--778}, year={2016}
}

@inproceedings{schoeffmann2018cataracts,
  title={{Cataract-101}: video dataset of 101 cataract surgeries},
  author={Schoeffmann, Klaus and Taschwer, Mario and Sarny, Stephanie and M{\"u}nzer, Bernd and Primus, Manfred J{\"u}rgen and Putzgruber, Doris},
  booktitle={Proceedings of the 9th ACM Multimedia Systems Conference (MMSys)},
  pages={421--425}, year={2018},
  doi={10.1145/3204949.3208137}
}

@article{ayobi2024grasp,
  title={Pixel-wise recognition for holistic surgical scene understanding},
  author={Ayobi, Nicol{\'a}s and Rodr{\'\i}guez, Santiago and P{\'e}rez, Alejandra and Hern{\'a}ndez, Isabela and Aparicio, Nicol{\'a}s and Dessevres, Eug{\'e}nie and Pe{\~n}a, Sebasti{\'a}n and Santander, Jessica and Caicedo, Juan Ignacio and Fern{\'a}ndez, Nicol{\'a}s and Arbel{\'a}ez, Pablo},
  journal={Medical Image Analysis},
  year={2024},
  note={arXiv preprint arXiv:2401.11174}
}

@article{ding2023uatd,
  title={Less is more: Surgical phase recognition from timestamp supervision},
  author={Ding, Xinpeng and Yan, Xinjian and Wang, Zixun and Zhao, Wei and Zhuang, Jian and Xu, Xiaowei and Li, Xiaomeng},
  journal={IEEE Transactions on Medical Imaging},
  volume={42}, number={6}, pages={1897--1910}, year={2023}
}

@inproceedings{yi2022note2e,
  title={Not End-to-End: Explore Multi-Stage Architecture for Online Surgical Phase Recognition},
  author={Yi, Fangqiu and Yang, Yanfeng and Jiang, Tingting},
  booktitle={ACCV}, year={2022}
}

@misc{dylan2017heico,
  title={Surgical workflow analysis in the {SensorOR} ({EndoVis} 2017 sub-challenge participant)},
  author={{Dylan team}},
  year={2017},
  howpublished={\url{https://endovissub2017-workflow.grand-challenge.org/}}
}

@misc{andrei2017heico,
  title={Surgical workflow analysis in the {SensorOR} ({EndoVis} 2017 sub-challenge participant)},
  author={{Andrei team}},
  year={2017},
  howpublished={\url{https://endovissub2017-workflow.grand-challenge.org/}}
}

@misc{robin2017heico,
  title={Surgical workflow analysis in the {SensorOR} ({EndoVis} 2017 sub-challenge participant)},
  author={{Robin team}},
  year={2017},
  howpublished={\url{https://endovissub2017-workflow.grand-challenge.org/}}
}

@article{bodenstedt2017heico,
  title={Unsupervised temporal context learning using convolutional neural networks for laparoscopic workflow analysis},
  author={Bodenstedt, Sebastian and Wagner, Martin and Katic, Darko and Mietkowski, Patrick and Mayer, Benjamin and Kenngott, Hannes and M{\"u}ller-Stich, Beat and Dillmann, R{\"u}diger and Speidel, Stefanie},
  journal={arXiv preprint arXiv:1702.03684}, year={2017}
}

@inproceedings{qi2019cataract,
  title={A deep architecture for surgical workflow recognition with edge information},
  author={Qi, Bo and Qin, Xianpeng and Liu, Jian and Xu, Yu and Chen, Yan},
  booktitle={IEEE International Conference on Bioinformatics and Biomedicine (BIBM)},
  pages={1358--1364}, year={2019}
}

@inproceedings{he2022activity,
  title={An empirical study on activity recognition in long surgical videos},
  author={He, Zhuohong and Mottaghi, Ali and Sharghi, Aidean and Jamal, Muhammad Abdullah and Mohareri, Omid},
  booktitle={Machine Learning for Health (ML4H)},
  pages={356--372}, year={2022}
}

@article{rcnest,
  title={Against spatial--temporal discrepancy: contrastive learning-based network for surgical workflow recognition},
  author={Xia, Tong and Jia, Fucang},
  journal={International Journal of Computer Assisted Radiology and Surgery},
  volume={16}, number={5}, pages={839--848}, year={2021}
}

@inproceedings{wang2021oadtr,
  title={OadTR: Online Action Detection with Transformers},
  author={Wang, Xiang and Zhang, Shiwei and Qing, Zhiwu and Shao, Yuanjie and Zuo, Zhengrong and Gao, Changxin and Sang, Nong},
  booktitle={Proceedings of the IEEE/CVF International Conference on Computer Vision (ICCV)},
  pages={7565--7575},
  year={2021}
}

@inproceedings{xu2021lstr,
  title={Long Short-Term Transformer for Online Action Detection},
  author={Xu, Mingze and Xiong, Yuanjun and Chen, Hao and Li, Xinyu and Xia, Wei and Tu, Zhuowen and Soatto, Stefano},
  booktitle={Advances in Neural Information Processing Systems (NeurIPS)},
  volume={34},
  pages={1086--1099},
  year={2021}
}

@ARTICLE{cmanet2025,
  author={Fan, Wenpei and Wang, Yaonan and Liu, Licheng and Zeng, Jiayi and Liu, Min},
  journal={IEEE Transactions on Circuits and Systems for Video Technology}, 
  title={CMANet: A TCN-RMamba-Attention Network for Surgical Phase Online Recognition}, 
  year={2026},
  volume={36},
  number={2},
  pages={1460-1472},
  doi={10.1109/TCSVT.2025.3599391}}

@article{zhang2024sprmamba,
  title={SPRMamba: surgical phase recognition for endoscopic submucosal dissection with mamba},
  author={Zhang, Xiangning and Zhang, Qingwei and Chen, Jinnan and Zhou, Chengfeng and Wang, Yaqi and Zhang, Zhengjie and Li, Xiaobo and Qian, Dahong},
  journal={IEEE Sensors Journal},
  year={2026},
  publisher={IEEE}
}

@inproceedings{endomamba2025,
  title={EndoMamba: an efficient foundation model for endoscopic videos via hierarchical pre-training},
  author={Tian, Qingyao and Liao, Huai and Huang, Xinyan and Yang, Bingyu and Lei, Dongdong and Ourselin, Sebastien and Liu, Hongbin},
  booktitle={International Conference on Medical Image Computing and Computer-Assisted Intervention},
  pages={224--234},
  year={2025},
  organization={Springer}
}

@inproceedings{lahoti2026mamba3,
  title={Mamba-3: Improved sequence modeling using state space principles},
  author={Lahoti, Aakash Sunil and Li, Kevin and Chen, Berlin and Wang, Caitlin and Bick, Aviv and Kolter, Zico and Dao, Tri and Gu, Albert},
  booktitle={International Conference on Learning Representations},
  volume={2026},
  pages={86173--86202},
  year={2026}
}

@inproceedings{orvieto2023lru,
  title={Resurrecting recurrent neural networks for long sequences},
  author={Orvieto, Antonio and Smith, Samuel L and Gu, Albert and Fernando, Anushan and Gulcehre, Caglar and Pascanu, Razvan and De, Soham},
  booktitle={International conference on machine learning},
  pages={26670--26698},
  year={2023},
  organization={PMLR}
}

@misc{yang2025gateddeltanet,
      title={Gated Delta Networks: Improving Mamba2 with Delta Rule}, 
      author={Songlin Yang and Jan Kautz and Ali Hatamizadeh},
      year={2025},
      eprint={2412.06464},
      archivePrefix={arXiv},
      primaryClass={cs.CL},
      url={https://arxiv.org/abs/2412.06464}, 
}

\clearpage
\appendix
\section{Preliminaries and notation}
\label{app:prelim}

\paragraph{Continuous-time SSM and zero-order hold.} The slow path
discretizes a continuous-time linear SSM,
\begin{equation}
\frac{\mathrm dh(t)}{\mathrm dt} = A\,h(t) + B(t)\,x(t),
\qquad y(t) = C(t)\,h(t),
\label{eq:ct_ssm}
\end{equation}
with $A$ constrained for stability. Holding the input constant over a
step of length $\Delta$ integrates \eqref{eq:ct_ssm} exactly and gives
\begin{equation}
h_n = \bar A\,h_{n-1} + \bar B\,x_n,
\qquad \bar A = \exp(\Delta A), \quad \bar B \approx \Delta B .
\label{eq:zoh}
\end{equation}
Since $A$ is a per-head scalar, $\bar A$ is an element-wise decay, and
$\Delta$ is the only quantity setting how much of $h_{n-1}$ survives a
step. \S\ref{sec:slowpath} reads $\lambda$ as a change of the variable
being integrated, which is why the same zero-order hold applies with
$\Delta$ replaced by $\alpha_n\Delta$.

\paragraph{Notation.} The inner dimension splits into $H$ heads of
width $P$, so $\dinner = H\,P$, and a clip of $L$ frames into
$n_c = L/\Cchunk$ chunks of size $\Cchunk$. The carried state is one
tensor in $\mathbb{R}^{H\times P\times N}$ per clip boundary, and it
is this tensor that state regramming rotates (\S\ref{sec:tstate}).

\paragraph{Two properties we rely on.} First, the chunked training
matmul and the frame-by-frame inference recurrence compute the same
function, so we train on chunked clips and deploy per-frame with
identical weights. Second, a per-head scalar $\bar A$ commutes with
any orthogonal matrix acting on the state, so decay and rotation may
be applied in either order; this is what lets a boundary rotation be
inserted without disturbing the exponential decay, and what keeps the
result analyzable in the SSD matrix framework
(Appendix~\ref{app:matrixview}).

\section{What intensity modulation buys}
\label{app:lambda}

\subsection{Setup}
\label{app:lam_setup}

By \eqref{eq:warpdisc} the slow path discretizes with
$\bar A_n=\exp(\alpha_n\Delta A)$ and
$\bar B_n\approx\alpha_n\Delta B_n$, where
$\alpha_n=1+\lambda_n\in[1,2]$. Writing $a:=|A|\Delta>0$ (recall
$A<0$) gives $\bar A_n=e^{-a\alpha_n}$, and the fraction of the input
written at step $s$ that survives at step $n>s$ is the product of the
intervening decays:
\begin{equation}
R(s,n):=\prod_{k=s+1}^{n}\bar A_k
=\underbrace{e^{-a(n-s)}}_{\text{decay by age}}\cdot
\underbrace{e^{-a\Lambda(s,n)}}_{\text{extra, from }\lambda},
\qquad
\Lambda(s,n):=\sum_{k=s+1}^{n}\lambda_k .
\label{eq:lam_kernel}
\end{equation}
Two readings of \eqref{eq:lam_kernel} are used repeatedly. Since
$\lambda\ge0$, the second factor never exceeds one: $\lambda$ can only
add forgetting. And what it adds depends on the \emph{total amount} of
$\lambda$ accumulated between $s$ and $n$, never on how high $\lambda$
rose at any single frame. Note also that $\alpha_n$ multiplies the input gain
$\bar B_n\approx\alpha_n\Delta B_n$, not only the decay.

For tractability we hold $\Delta$ constant throughout this appendix, so
that $\alpha_n$ is the only source of per-frame variation in the decay
and $a$ is a single number. Mamba2's selective step is itself
input-dependent; \S\ref{app:lam_caveats} returns to what that means
for the argument.

\paragraph{Part I: why a separate control is needed.}

\subsection{A burst discards a fixed proportion, once per boundary}
\label{app:lam_what}

\emph{Each boundary crossed multiplies what remains by a fixed factor,
independent of when it occurred or how long the segments were.}

Let $\lambda$ rise to a burst around each boundary $b_1<b_2<\dots$ and
return to a floor $\lambda_{\mathrm{floor}}$ in between. Write $M$ for
the amount of $\lambda$ in one burst above that floor, and $N(s,n)$
for how many boundaries fall in $(s,n]$. Then
$\Lambda(s,n)=\lambda_{\mathrm{floor}}(n-s)+M\,N(s,n)$, and
\eqref{eq:lam_kernel} becomes
\begin{equation}
R(s,n)=\underbrace{e^{-a_{\mathrm{eff}}(n-s)}}_{\text{age}}
\cdot\underbrace{\gamma^{\,N(s,n)}}_{\text{boundaries}},
\qquad
a_{\mathrm{eff}}:=a(1+\lambda_{\mathrm{floor}}),
\quad
\gamma:=e^{-aM}.
\label{eq:lam_geo}
\end{equation}
The first factor is ordinary decay by age. The second is new: each
boundary crossed multiplies what remains by $\gamma$, a fixed
survival fraction that does not depend on when the crossing occurred
or on how long the segments on either side were. The strength of the
whole mechanism is one number, the burst size $M$, through
$\gamma=e^{-aM}$.

Two things follow immediately. A component of $\lambda$ that is
present everywhere contributes only to $a_{\mathrm{eff}}$, which is
indistinguishable from running the unmodulated model with
$\Delta\mapsto(1+\lambda_{\mathrm{floor}})\Delta$; in particular
$\lambda\equiv1$ is \emph{exactly} the unmodulated model at twice the
step. Holding $\lambda$ high everywhere is the same as never raising
it, and the mechanism lives entirely in the contrast between burst and
background. Second, if boundaries fall closer together than the bursts
are wide, adjacent bursts overlap, $\lambda$ never returns to its
floor, and by the same argument the mechanism degenerates into a
rescaled step. This is the condition $W\ll\mu$ recorded in
\S\ref{app:lam_disp}.

\subsection{A fixed step cannot forget without also forgetting the present}
\label{app:lam_cost}

\emph{A fixed step decays by age alone, so it cannot suppress the old
segment without suppressing the new one at the same rate; a burst
can.}

Compare what remains at step $n$ of two inputs written the same number
of steps earlier, $\ell=n-s$: one inside the current segment, one
before the last boundary.
Under a fixed step both are multiplied by $e^{-a\ell}$, so
\begin{equation}
S_{\lambda\equiv0}:=\frac{R_{\text{within}}(\ell)}{R_{\text{across}}(\ell)}\equiv1
\qquad\text{for every }a,
\label{eq:lam_sel_fixed}
\end{equation}
whereas from \eqref{eq:lam_geo} the modulated kernel gives
\begin{equation}
S=\gamma^{-1}=e^{aM}.
\label{eq:lam_sel}
\end{equation}
Equation~\eqref{eq:lam_sel_fixed} is an identity: a kernel that depends
only on how many steps have elapsed cannot treat the two cases
differently, whatever its rate. The price is quantifiable. To suppress
pre-boundary content by the same factor $\gamma=e^{-aM}$ at age $u$
using the rate alone, we need
$e^{-2a'u}=\gamma^{2}e^{-2au}=e^{-2a(u+M)}$; matching exponents gives
\begin{equation}
a' \;=\; a\,\frac{u+M}{u} \;=\; a + \frac{aM}{u}.
\label{eq:lam_aprime}
\end{equation}
That larger rate also governs how much of its own content the current
segment retains; with $E_\infty(a)=\beta^2/(1-e^{-2a})\approx\beta^2/2a$
for incoherent inputs of per-frame energy $\beta^2$, so that
$E_\infty\propto1/a$,
\begin{equation}
\frac{E_\infty(a')}{E_\infty(a)}\approx\frac{a}{a'}=\frac{u}{u+M}.
\label{eq:lam_cost_eq}
\end{equation}
The fixed step buys suppression and pays for it with capacity, in
fixed proportion, because one parameter serves both. A burst attaches its
suppression to the boundary rather than to the age, achieving
\eqref{eq:lam_sel} while \eqref{eq:lam_cost_eq} stays at one.
Decomposing the state shortly after a boundary,
\begin{equation}
h_{b+u}=\underbrace{R(b,b+u)\,h_b}_{\text{carried over}}
+\underbrace{\sum_{k>b}\bar B_k\,R(k,b+u)\,x_k}_{\text{written since}},
\label{eq:lam_split}
\end{equation}
only the first term crosses the burst and acquires $\gamma$. The second
carries gain $\bar B_k=\alpha_k\Delta B_k$ with $\alpha_k$ up to $2$,
so
\begin{equation}
\frac{E_{\text{written since}}}{E_{\text{carried over}}}
\;\propto\;\underbrace{\gamma^{-2}}_{\text{old content cleared}}
\cdot\underbrace{\overline{\alpha^{2}}}_{\text{new content written harder}},
\label{eq:lam_snr}
\end{equation}
with $\overline{\alpha^{2}}$ the mean squared gain over the burst,
approaching $4$ at its peak. Both factors come from the same $\alpha$,
which shortens what the state keeps and strengthens what it takes in
over the same frames. A fixed step has no such pairing: raising $a$
suppresses old content, leaves $\bar B$ untouched, and pays
\eqref{eq:lam_cost_eq} on the current segment. This coupling, not forgetting as such, is what
$\lambda$ contributes.

\subsection{The separation is worth most when segments differ in length}
\label{app:lam_disp}

\emph{What the mechanism recovers scales with the ratio between the
longest and shortest segments, and vanishes when they are equal.}

Fix a discard threshold $\theta$: by the end of a segment, content
from before it began should have decayed to at most $\theta$. A fixed
step decays only with elapsed time, so crossing segment $m$ of length
$L_m$ supplies $e^{-aL_m}$ and the requirement $e^{-aL_m}\le\theta$
must hold for every segment, giving
\begin{equation}
a\;\ge\;\frac{\log(1/\theta)}{L_{\min}},
\qquad L_{\min}:=\min_m L_m .
\label{eq:lam_amin}
\end{equation}
The shortest segment sets the bound, because it offers the least time
in which to forget. Retention pulls the other way: the capacity a
segment has for its own content is $E_\infty(a)\approx\beta^2/2a$,
increasing as $a$ decreases. The best a fixed step can do is therefore
to sit exactly at \eqref{eq:lam_amin}, and its capacity is
\begin{equation}
\frac{1}{a^{\star}_{\mathrm{fixed}}}=\frac{L_{\min}}{\log(1/\theta)} .
\label{eq:lam_capfixed}
\end{equation}
Every segment (including one an order of magnitude longer) is held
to a horizon fixed by the shortest.

A burst changes which quantity carries the requirement. By
\eqref{eq:lam_geo} a crossing contributes the factor
$\gamma=e^{-aM}$ regardless of the length of the segment crossed, so
the same threshold reads
\begin{equation}
\gamma\le\theta
\quad\Longleftrightarrow\quad
aM\;\ge\;\log(1/\theta),
\label{eq:lam_aMbound}
\end{equation}
in which $L_m$ does not appear. The constraint has moved from $a$ to
$M$, leaving $a$ free to be set by retention alone, by the longest
horizon the task requires, $a^{\star}_{\lambda}\approx1/L_{\max}$
rather than by $L_{\min}$. Comparing capacities,
\begin{equation}
\frac{1/a^{\star}_{\lambda}}{1/a^{\star}_{\mathrm{fixed}}}
=\log(1/\theta)\,\cdot\,\frac{L_{\max}}{L_{\min}} .
\label{eq:lam_ratio}
\end{equation}

Equation~\eqref{eq:lam_ratio} is the answer to when the mechanism
matters: the recoverable capacity scales with the spread of segment
lengths, through their ratio. If all segments have the same length
then $L_{\min}=L_{\max}$ and the two constraints coincide: a single
$a=\log(1/\theta)/L$ discards exactly enough at each boundary while
retaining within segments, so elapsed time is a faithful proxy for
boundary crossing and $\lambda$ has nothing to add. The same follows
from the kernel: with $L_m\equiv L$, $\gamma^{N(\ell)}$ is itself
exponential in $\ell$ and a fixed step reproduces the modulated kernel
to within $\gamma^{\pm1}$. As the lengths spread apart, age stops
determining whether a boundary has been crossed, and the fixed step
must choose which end of the distribution to serve. Since $L_{\min}$
is sensitive to a single outlying segment, we use a low quantile $L_q$
in its place, so that \eqref{eq:lam_ratio} reads $L_{\max}/L_q$.

Two further requirements bracket the regime: the horizon
must span a segment, $a\mu\lesssim1$ with $\mu$ the mean length, or
nothing survives to be suppressed; and bursts must not overlap,
$W\ll\mu$, or $\lambda$ acquires a floor and degenerates
(\S\ref{app:lam_what}). With $M\le W$ (\S\ref{app:lam_range}) and
\eqref{eq:lam_aMbound}, the usable window is
$\log(1/\theta)\lesssim aW\ll a\mu\lesssim1$: narrow, which predicts a
consistent rather than dramatic effect, in line with the $-1.6$\,pp
strict-Jaccard drop when $\lambda$ is removed
(Table~\ref{tab:component_ablation}), and which pins the burst width
to a band scaling with segment length rather than wall-clock time.
Both $W/\mu$ and $L_{\max}/L_q$ follow from segment annotations alone,
so whether a new task lies in this regime is checkable before
training.

\paragraph{Part II: how the control is configured.}

\subsection{Why the control is capped at \texorpdfstring{$\alpha=2$}{alpha=2}}
\label{app:lam_range}

\emph{A burst suppresses by lasting rather than by peaking, so capping
its height costs little.}

At the peak, $\alpha_n=2$ and $\bar A_n=e^{-2a}=(\bar A^{(0)})^2$: the
per-frame survival is squared and the local half-life halved. That is
a modest change on its own, with the within-phase plateau measured at
$\bar A\approx0.92$ (\S\ref{sec:qualitative}), so $a\approx0.083$, a
peak frame decays at $0.846$. The suppression that matters comes from
$M$, and since $\lambda\le1$ a burst of width $W$ has $M\le W$, so
\begin{equation}
S=e^{aM}\le e^{aW}.
\label{eq:lam_bound}
\end{equation}
Halving the surviving pre-boundary state ($S=2$) needs
$M=\log2/a\approx8.3$, about eight frames of full intensity, eight
seconds at the $1$\,fps sampling of \S\ref{sec:setup}. The cap is
thus not a limitation on how much can be forgotten but a requirement
that forgetting be spread over frames: it forbids
$\alpha\to\infty$, which would drive $\bar A\to0$ and erase the state
in one step. The lower end $\alpha\ge1$ forbids deceleration, which
would push $\bar A\to1$ and freeze it, and makes $\lambda\to0$ recover
the unmodulated model exactly: a nested baseline the block cannot
underperform given capacity.

\subsection{Why the burst is asymmetric}
\label{app:lam_asym}

\emph{The same amount of forgetting costs less after a boundary than
before it.}

The target of \S\ref{sec:loss} rises with width $\sigma_L$ before the
boundary and decays with width $\sigma_R$ after it. With $u=t-b$,
\begin{equation}
\lambda(u)=
\begin{cases}
\exp(-u^{2}/2\sigma_L^{2}), & u<0,\\[3pt]
\exp(-u^{2}/2\sigma_R^{2}), & u\ge0,
\end{cases}
\qquad
M_{L,R}=\!\int\!\lambda=\sqrt{\tfrac{\pi}{2}}\,\sigma_{L,R},
\quad
M=\sqrt{\tfrac{\pi}{2}}(\sigma_L+\sigma_R).
\label{eq:lam_target}
\end{equation}
By \eqref{eq:lam_sel} the suppression depends on the \emph{sum} only:
$(\sigma_L,\sigma_R)=(1,9)$ and $(5,5)$ discard equally. The split
decides not how much is forgotten but which segment pays for it. Three
independent arguments put the mass on the right.

\paragraph{The left side strikes a full state.} A segment accumulates
$E(u)\approx\beta^2\min(u,1/2a)$ of its own content, so before the
boundary the outgoing segment is saturated at $\beta^2/2a$ while after
it the incoming one is nearly empty, $E(u)\approx\beta^2u$. Mass
placed where the state holds more destroys more; with
$\mathrm dM=\lambda(u)\,\mathrm du$, integrating
$\mathrm dD\approx2a\,E(u)\,\mathrm dM$ over each side gives
\begin{equation}
\begin{aligned}
D_L &\approx \int_{-\infty}^{0}\!2a\,\frac{\beta^{2}}{2a}\,e^{-u^{2}/2\sigma_L^{2}}\,\mathrm du
= \beta^{2}\sqrt{\tfrac{\pi}{2}}\,\sigma_L
&&(\text{linear}),\\[2pt]
D_R &\approx \int_{0}^{\infty}\!2a\,\beta^{2}u\;e^{-u^{2}/2\sigma_R^{2}}\,\mathrm du
= 2a\beta^{2}\sigma_R^{2}
&&(\text{quadratic, scaled by }2a),
\end{aligned}
\label{eq:lam_DLDR}
\end{equation}
using $\int_0^\infty u\,e^{-u^2/2\sigma^2}\mathrm du=\sigma^2$.
Comparing marginal costs at fixed $\sigma_L+\sigma_R$ favours the
right whenever $\sigma_R<0.31/a$. Damage on the left is also
unrecoverable, since the segment it harms ends at $b$, whereas a
segment harmed just after $b$ continues and refills.

\paragraph{Only the right side benefits from the write gain.} The
amplification in \eqref{eq:lam_snr} acts on whatever is written while
$\lambda$ is elevated, and the two sides write different things: to
the right, the incoming segment, which the gain helps establish; to
the left, the final frames of a segment that is discarded moments
later. Requiring the net effect on the incoming segment to be an
amplification, $2e^{-aM_R}>1$, bounds
\begin{equation}
\sigma_R<\frac{\log2}{a\sqrt{\pi/2}}=\frac{0.55}{a},
\label{eq:lam_sigR}
\end{equation}
close to the $\sigma_R\lesssim1/2a$ at which $E(u)$ saturates and the
quadratic advantage in \eqref{eq:lam_DLDR} disappears; two arguments,
the same scale.

\paragraph{Only the right side is observable.} $\lambda$ is predicted
causally (\S\ref{sec:slowpath}), so mass before the boundary asks the
model to raise it before the evidence for a transition has appeared,
while mass after it asks the model to recognize evidence already
present. The same causality fixes the target's shape: if the true
transition is $b+\epsilon$ with $\epsilon\sim p$, the pointwise-optimal
target is $\mathbb E_\epsilon[\delta(u-\epsilon)]=p(u)$, so a soft
target is correct under boundary uncertainty rather than an
approximation to a hard one, with $\sigma$ its scale and $p$ skewed
right. All three arguments give
$\sigma_L\ll\sigma_R$, bounded above by \eqref{eq:lam_sigR}.

\subsection{What is established, and what is not}
\label{app:lam_caveats}

Equations~\eqref{eq:lam_kernel}--\eqref{eq:lam_sel} are identities. The split of
retention into age and boundary factors, the degeneracy of a constant
$\lambda$, the inability of a fixed step to distinguish pre-boundary
from within-segment content, and the bound $S\le e^{aW}$ all hold
exactly, as does the uniform-length case of
\S\ref{app:lam_disp} with error bounded by $\gamma^{\pm1}$.

The quantitative statements do not.
Equations~\eqref{eq:lam_cost_eq}, \eqref{eq:lam_snr},
\eqref{eq:lam_capfixed}, \eqref{eq:lam_ratio} and
\eqref{eq:lam_DLDR} assume mutually incoherent inputs of constant
per-frame energy and measure damage as energy loss, whereas the
read-out $C_t$ selects directions; \eqref{eq:lam_snr} further reads
the gain as a uniform scaling, exact only for
$\bar B_n\approx\alpha_n\Delta B_n$. Equation~\eqref{eq:lam_ratio}
treats the discard threshold $\theta$ as a hard requirement on every
segment and the retention horizon as set by $L_{\max}$; both are
idealizations of what the loss actually enforces, and the resulting
ratio should be read as the scale of the shortfall rather than as a
predicted speedup. More fundamentally, the whole
appendix accounts for the mechanism in units of retained state
capacity rather than task loss. That recovering this capacity improves
predictions is measured by the ablation of \S\ref{sec:ablation}, not
proved here. Nor does the appendix isolate $\lambda$ as the only way to obtain this
schedule. Holding $\Delta$ fixed is what lets its effect be read off,
and a selective $\Delta$ could in principle vary in the same way. In
the model itself $\lambda$ multiplies that input-dependent step rather
than replacing it, and what the mechanism contributes is the
supervision that ties forgetting to event structure
(\S\ref{sec:rel_stepping}).

\section{Matrix view of state regramming}
\label{app:matrixview}

This appendix proves Proposition~\ref{prop:preserve} and derives the
matrix decomposition it summarizes, following the notation of
\S\ref{sec:method}.

\begin{proposition}[Structure preservation]
\label{prop:preserve}
Let $Z^{(c)}\!\in O(N)$ be applied to the carried state at each chunk
boundary of the SSD scan with per-head scalar transition. Then (i) the
induced sequence map remains $N$-semiseparable, so per-frame inference
cost remains $O(d)$; and (ii) the rotation commutes with the scalar
decay, so decay-then-rotate and rotate-then-decay realize the same
map. Cross-chunk blocks take the form
\begin{equation}
M_{cc'} = L \odot \bigl(B\,\Zstate_{c',c}\,C^{\top}\bigr),
\qquad
\Zstate_{c',c} = \textstyle\prod_{k=c'}^{c-1} Z^{(k)},
\label{eq:crossblock}
\end{equation}
with $\Zstate_{c',c}$ the ordered product of boundary rotations
between chunks $c'$ and $c$.
\end{proposition}

The remainder of this appendix proves the proposition and reads off
what \eqref{eq:crossblock} implies for the kernel.

\subsection{Setup}
\label{app:setup}

We follow the notation of \S\ref{sec:method}. Fix a single head with per-head scalar $A \in \mathbb{R}_{<0}$ and per-head width $P = 1$; the general case is recovered by stacking. Frames are indexed by $n$, and the discrete decay at frame $n$ is $\bar a_n = \exp(\Delta_n A)$. Define the cumulative decay
\begin{equation}
a(t{:}s) \;:=\; \prod_{n = s+1}^{t} \bar a_n,
\qquad a(t{:}t) = 1.
\end{equation}
A sequence of length $T$ is partitioned into chunks of size $\Cchunk$, indexed by $c \in \{0, 1, 2, \dots\}$. Within chunk $c$, the per-frame selective vectors $B^{(c)}_s \in \mathbb{R}^{N}$ and $C^{(c)}_t \in \mathbb{R}^{N}$ are computed from the chunk's input. The boundary rotation $Z^{(c)} \in O(N)$ applied at the end of chunk $c$ is the Cayley map of an input-conditioned skew-symmetric matrix, as defined in \S\ref{sec:tstate}.

\paragraph{Conventions.}
Following the boundary update of \S\ref{sec:tstate}, we adopt the row-vector convention throughout this appendix. The recurrence is
\begin{equation}
h_t \;=\; \bar a_t\,h_{t-1} \;+\; x_t\,B_t^{\top},
\qquad y_t \;=\; h_t\,C_t,
\end{equation}
state regramming applies $h^{(c)} \leftarrow h^{(c)}\,Z^{(c)}$ at chunk boundaries, and $y_t \in \mathbb{R}$ is a scalar under $P = 1$. Transposes appearing on $B$ and $C$ reflect this row-vector setup.

\subsection{Intra-chunk block (vanilla SSD)}
\label{app:intrachunk}

\emph{Inside a chunk nothing changes, and the vanilla SSD block is
recovered exactly.}

Inside chunk $c$, no rotation is applied. Unrolling the recurrence from $h_{-1}^{(c)} = 0$ gives, for $t, s$ both in chunk $c$,
\begin{equation}
h_t^{(c)} \;=\; \sum_{u = 0}^{t} a(t{:}u)\,x_u\,(B^{(c)}_u)^{\top},
\qquad
y_t^{(c)} \;=\; h_t^{(c)}\,C^{(c)}_t \;=\; \sum_{u = 0}^{t} a(t{:}u)\,\bigl((B^{(c)}_u)^{\top} C^{(c)}_t\bigr)\,x_u.
\end{equation}
Writing $y = M_{cc}\,x$ for the chunk's input--output map, the intra-chunk block has entries
\begin{equation}
\boxed{\;M_{cc}[t, s] \;=\; a(t{:}s)\,(B^{(c)}_s)^{\top}\,C^{(c)}_t\;}
\label{eq:intra}
\end{equation}
This is the standard SSD form: a $1$-semiseparable kernel $L^{(c,c)}$ with $L^{(c,c)}[t,s] = a(t{:}s)$, element-wise multiplied with the rank-$N$ outer product $C^{(c)}\,(B^{(c)})^{\top}$ (so that $\bigl(C^{(c)}(B^{(c)})^{\top}\bigr)[t,s] = C^{(c)}_t \cdot B^{(c)}_s = (B^{(c)}_s)^{\top} C^{(c)}_t$, matching \eqref{eq:intra}). State regramming does not modify this block.

\subsection{Cross-chunk block with one boundary rotation}
\label{app:onerotation}

\emph{One boundary rotation turns the diagonal pairing of written and
read dimensions into a dense one.}

Consider two adjacent chunks $c'$ and $c' + 1$, and let $h_{\Cchunk-1}^{(c')}$ denote the final hidden state of chunk $c'$. The rotation $Z^{(c')}$ is applied at the boundary, so the next chunk receives
\begin{equation}
h_0^{(c'+1)} \;:=\; h_{\Cchunk-1}^{(c')}\,Z^{(c')}.
\end{equation}
For $t$ in chunk $c' + 1$ and $s$ in chunk $c'$, we trace the contribution of $x_s$ to $y_t$. From \S\ref{app:intrachunk},
\begin{equation}
h_{\Cchunk-1}^{(c')} \;\supset\; a(\Cchunk{-}1{:}s)\,x_s\,(B^{(c')}_s)^{\top},
\end{equation}
where ``$\supset$'' denotes the $x_s$-contribution. After rotation,
\begin{equation}
h_0^{(c'+1)} \;\supset\; a(\Cchunk{-}1{:}s)\,x_s\,(B^{(c')}_s)^{\top}\,Z^{(c')}.
\end{equation}
Within chunk $c' + 1$, this initial-state contribution decays by $a(t{:}\Cchunk{-}1)$ before being read by $C^{(c'+1)}_t$. Composing the decays as $a(t{:}\Cchunk{-}1)\,a(\Cchunk{-}1{:}s) = a(t{:}s)$ and identifying the coefficient of $x_s$,
\begin{equation}
\boxed{\;M_{(c'+1)\,c'}[t, s] \;=\; a(t{:}s)\,(B^{(c')}_s)^{\top}\,Z^{(c')}\,C^{(c'+1)}_t\;}
\label{eq:cross_one}
\end{equation}
Compared with the vanilla cross-chunk form $a(t{:}s)\,(B_s)^{\top} C_t$, the only change is the orthogonal factor $Z^{(c')}$ inserted between the chunk-$c'$ write and the chunk-$(c'+1)$ read.

\paragraph{Element-wise reading.}
Expanding the bilinear form in \eqref{eq:cross_one},
\begin{equation}
(B^{(c')}_s)^{\top}\,Z^{(c')}\,C^{(c'+1)}_t \;=\; \sum_{i = 0}^{N-1}\sum_{j = 0}^{N-1} B^{(c')}_{s,i}\,z^{(c')}_{ij}\,C^{(c'+1)}_{t,j}.
\label{eq:elementwise}
\end{equation}
Contrast with the vanilla case $B_s^{\top} C_t = \sum_{i} B_{s,i}\,C_{t,i}$, which pairs only matched dimensions. State regramming replaces this diagonal pairing with a content-dependent dense pairing weighted by $z^{(c')}_{ij}$, so that any state dimension $i$ written by $B^{(c')}_s$ can be read by any state dimension $j$ of the next chunk's $C^{(c'+1)}_t$. The write and read coordinate systems, tied to a common axis in vanilla SSD, are thereby decoupled, while the SSD scan structure of \eqref{eq:intra} is left intact.

\subsection{Composition across multiple chunk boundaries}
\label{app:composition}

\emph{Rotations accumulate in order, one per boundary crossed, and
each stays in the map thereafter.}

Carrying the same unrolling across $k$ boundaries, the contribution of $x_s$ in chunk $c'$ to $y_t$ in chunk $c' + k$ accumulates one rotation per boundary crossed. Define
\begin{equation}
\mathcal{Z}_{c',c} \;:=\; Z^{(c')}\,Z^{(c'+1)}\cdots Z^{(c-1)} \;=\; \prod_{j = c'}^{c - 1} Z^{(j)}
\qquad (c > c'),
\qquad \mathcal{Z}_{c,c} := I.
\end{equation}
The general block of the full transfer matrix is then
\begin{equation}
\boxed{\;M_{cc'}[t, s] \;=\; a(t{:}s)\,(B^{(c')}_s)^{\top}\,\mathcal{Z}_{c',c}\,C^{(c)}_t\;}
\qquad (c \ge c'),
\label{eq:general}
\end{equation}
with $M_{cc'} = 0$ for $c < c'$ (causal).

\paragraph{Block structure of $M$.}
For $T = 4\Cchunk$ the full transfer matrix is block lower-triangular,
\begin{equation}
M \;=\; \begin{pmatrix}
M_{00} & 0 & 0 & 0\\
M_{10} & M_{11} & 0 & 0\\
M_{20} & M_{21} & M_{22} & 0\\
M_{30} & M_{31} & M_{32} & M_{33}
\end{pmatrix},
\end{equation}
with one boundary rotation accumulated per super-diagonal step away from the main diagonal. Table~\ref{tab:z_trail} lists the rotation product $\mathcal{Z}_{c',c}$ appearing in each block.

\begin{table}[htbp]
\centering
\caption{Rotation factor $\mathcal{Z}_{c',c}$ appearing in each block $M_{cc'}$ of the transfer matrix for $T = 4\Cchunk$. Diagonal blocks reduce to vanilla SSD; each super-diagonal step accumulates one boundary rotation.}
\label{tab:z_trail}
\small
\setlength{\tabcolsep}{8pt}
\begin{tabular}{c|cccc}
\toprule
$c \,\backslash\, c'$ & $0$ & $1$ & $2$ & $3$ \\
\midrule
$0$ & $I$              & $0$        & $0$    & $0$ \\
$1$ & $Z^{(0)}$        & $I$        & $0$    & $0$ \\
$2$ & $Z^{(0)}Z^{(1)}$ & $Z^{(1)}$  & $I$    & $0$ \\
$3$ & $Z^{(0)}Z^{(1)}Z^{(2)}$ & $Z^{(1)}Z^{(2)}$ & $Z^{(2)}$ & $I$ \\
\bottomrule
\end{tabular}
\end{table}

\paragraph{Orthogonality is closed under composition.}
A product of orthogonal matrices is orthogonal, so $\mathcal{Z}_{c',c} \in O(N)$ for every $(c', c)$. The carried state's norm is preserved exactly across an arbitrary number of chunk boundaries; state regramming does not accumulate amplification or attenuation. The geometric decay of long-horizon contributions is governed entirely by $a(t{:}s)$, exactly as in vanilla Mamba2, and what the rotation changes is the direction in which those contributions are held.

\paragraph{The ``lifetime'' of a single rotation.}
A given $Z^{(c)}$ appears in every block $M_{c''c'}$ with $c' \le c < c''$: once applied at the end of chunk $c$, it is permanently embedded in the propagation of every chunk-$c'$ (or earlier) input to every chunk-$c''$ (or later) output, rather than acting once and expiring.

\paragraph{What the vanilla kernel cannot express.}
Equation~\eqref{eq:general} is best read against its vanilla
counterpart, in which the kernel factorizes as
$a(t{:}s)\cdot(B^{(c')}_s)^{\top}C^{(c)}_t$. How much a past frame
contributes to a present one is fixed by the contents of those two
frames and by their separation alone, so nothing that happened in
between can influence it. Inserting $\mathcal{Z}_{c',c}$ breaks that
factorization, since the rotation product is a function of every chunk
between $s$ and $t$. Three consequences follow.

\emph{Collision.} Two past frames with similar content have similar
$B_s$ and are therefore written to nearly the same direction, where
they superpose; no read-out can then recover one without the other.
Under \eqref{eq:general} their effective write directions are
$B_{s}^{\top}\mathcal{Z}_{c(s),c}$ and
$B_{s'}^{\top}\mathcal{Z}_{c(s'),c}$, which differ whenever the
intervening chunks differ, so repeated content is separated by the
path taken between occurrences rather than by content alone.

\emph{Pairing.} In the vanilla kernel a state dimension can only be
read by the query component carrying its own index
(\S\ref{app:onerotation}), and that correspondence is fixed for the
whole stream; under state regramming it is re-chosen at every boundary
from the content of the chunk that just ended.

\emph{Where discrimination lives.} In the vanilla kernel the only
quantity separating two contributions of the same content is
$a(t{:}s)$, which is monotone in elapsed time and saturates, so once both decays
are small the two become indistinguishable in relative terms as well.
State regramming does not change the magnitude $a(t{:}s)$, since
long-horizon contributions decay geometrically exactly as before, but
it places the distinction in direction, which neither
saturates nor is tied to elapsed time. Contributions of comparable age
therefore remain separable, and separability no longer degrades with
the age of the pair. Finally, because orthogonal matrices do not
commute in general, $\mathcal{Z}_{c',c}$ reflects the trajectory of
intervening chunk contents rather than their unordered set.

\subsection{Read-out, SMA view, and N-semiseparability}
\label{app:smaview}

\emph{Written as a factor between query and key, the rotation leaves
the rank bound and the per-frame cost untouched.}

\paragraph{Effective read-out.}
Equation \eqref{eq:general} admits an algebraically equivalent reorganization absorbing the rotation into the read-out:
\begin{equation}
\widetilde{C}^{(c,c')}_t \;:=\; \mathcal{Z}_{c',c}\,C^{(c)}_t,
\qquad M_{cc'}[t, s] \;=\; a(t{:}s)\,(B^{(c')}_s)^{\top}\,\widetilde{C}^{(c,c')}_t.
\end{equation}
The effective read-out $\widetilde{C}^{(c,c')}_t$ is a path-dependent, history-aware projection of the original $C^{(c)}_t$. Although $C^{(c)}_t$ is computed only from chunk-$c$ input, the basis in which it reads a chunk-$c'$ memory is shaped by every intervening boundary through $\mathcal{Z}_{c',c}$. Componentwise, $\widetilde{C}_{t,i} = \sum_j (\mathcal{Z}_{c',c})_{ij}\,C^{(c)}_{t,j}$, so the coefficient retrieving the dimension that $B^{(c')}_{s,i}$ wrote is a mixture of \emph{all} components of $C^{(c)}_t$, against $C^{(c)}_{t,i}$ alone in vanilla SSD. This is the dense pairing of \eqref{eq:elementwise} read from the other side.

The read-out's rank is unchanged, since an orthogonal rotation preserves rank, so $\widetilde{C}^{(c,c')}_t$ remains a rank-$1$ projection per output channel. State regramming does not enlarge the read-out capacity within any single chunk; it only redirects what each chunk reads. This is consistent with the empirical observation in Appendix~\ref{app:hparam} that the rotation rank $r$ used to parameterize the skew-symmetric generator has only a mild effect on accuracy.

\paragraph{SMA view.}
Within the structured state-space duality of Mamba2, the standard SMA form reads
\begin{equation}
Y_t \;=\; \sum_s L_{ts}\,Q_t\,K_s^{\top}\,V_s
\qquad \text{(standard SMA)},
\end{equation}
where $(Q, K, V)$ correspond to $(C, B, x)$ and $L_{ts} = a(t{:}s)$ on the lower triangle. State regramming generalizes this to
\begin{equation}
Y_t \;=\; \sum_s L_{ts}\,Q_t\,W_{c(t), c(s)}\,K_s^{\top}\,V_s,
\qquad W_{c(t), c(s)} \;=\; \mathcal{Z}_{c(s), c(t)}^{\top},
\label{eq:sma_general}
\end{equation}
where $c(\cdot)$ maps a frame index to its chunk. The new factor $W$ is a chunk-pair-dependent orthogonal matrix inserted between query and key, equal to the identity for same-chunk interactions and accumulating one rotation per boundary crossed. Only the query--key contraction is modified; the temporal kernel $L$ and the value path $V$ are untouched.

\paragraph{Four-factor block decomposition and N-semiseparability.}
The vanilla SSD off-diagonal block factorizes as $M_{cc'}^{\text{std}} = L^{(c,c')}\odot\bigl(B^{(c')}(C^{(c)})^{\top}\bigr)$, exposing two rank-controlling factors. State regramming yields the four-factor form
\begin{equation}
M_{cc'} \;=\; L^{(c,c')}\,\odot\,\Bigl(B^{(c')}\,\mathcal{Z}_{c',c}\,(C^{(c)})^{\top}\Bigr).
\end{equation}
The new factor $\mathcal{Z}_{c',c}$ is an $N \times N$ orthogonal matrix and therefore preserves the rank bound: $\operatorname{rank}(M_{cc'}) \le N$. State regramming retains the $N$-semiseparable structure of Mamba2's SSD form, and consequently its $O(d)$ per-frame inference cost. The only additional per-chunk operations are one $N \times N$ Cayley map and one $N \times N$ orthogonal multiply on the state, both amortized over $\Cchunk$ frames.

\medskip

Taken together, the analysis in this appendix shows that state regramming is a conservative extension of Mamba2's chunked SSD scan: the intra-chunk block, the chunk-granular state shape, the $N$-semiseparable rank bound, the $O(d)$ per-frame inference cost, and the geometric decay of long-horizon contributions are all preserved, while the basis in which carried information is read becomes content-dependent and path-dependent through the accumulated boundary rotations. What the extension adds is what the vanilla factorization excludes, a dependence of the kernel on what lay between two frames rather than on the two frames and how far apart they are. This is the claim of \S\ref{sec:tstate} in matrix form: the same content arriving in a different context is written along a different direction, so a read-out can retrieve one occurrence without the other.

\section{Relation to Mamba-3}
\label{app:mamba3}

Mamba-3~\citep{lahoti2026mamba3} also introduces rotations into an SSD
recurrence, and separately changes the discretization scheme. Both are
worth setting against what we do, since the mechanisms act on
different objects and answer different questions.

\paragraph{Why the rotation is there, and where.} The two rotations
answer different needs. Mamba-3 introduces one because real
eigenvalues can only decay: a state governed by them cannot cycle, so
tasks requiring periodic state, such as parity or modular arithmetic,
are out of reach, and complex transitions restore the rotational
dynamics those tasks need. Our rotation is not about periodicity at
all, but about where content is placed, and orthogonality is what
makes the re-placement lossless (\S\ref{sec:tstate}).

Mechanically, Mamba-3 makes the state transition complex. Its Proposition~2 writes the equivalent real
recurrence with a doubled state and transition
$\bar A_t = e^{\Delta_t A_t}R_t$, a scalar decay times a
block-diagonal rotation, so the state is rotated at every step. To
keep the SSD form, that rotation is not carried out inside the scan:
their Proposition~3 shows the same recurrence is obtained by passing
only the real part to the scan, as in Mamba-2, and applying the
rotation to the $B$ and $C$ projections as a rotary embedding, with
per-head angles scaled by $\Delta_t$ and accumulated over time.
Writing $R_{1:t}$ for that accumulation, the projections used at step
$t$ are $R_{1:t}B_t$ and $R_{1:t}C_t$. State regramming leaves $B$ and
$C$ untouched and multiplies the carried state at chunk boundaries
instead.

\paragraph{Both alter the interval factor, but not in the same way.}
Substituting the rotated projections into the read-out gives, for what
step $s$ leaves behind at step $t$,
\begin{equation}
(R_{1:s}B_s)^{\top}(R_{1:t}C_t)
= B_s^{\top}\,R_{s:t}\,C_t,
\qquad
R_{s:t}=\prod_{k=s+1}^{t}R_k,
\label{eq:m3_interval}
\end{equation}
so Mamba-3's kernel, like ours, carries a factor determined by the
steps between $s$ and $t$ rather than by the two endpoints alone. The
difference is what that factor can encode. Each $R_k$ is
block-diagonal, a plane rotation in each channel pair, and rotations
sharing a common set of planes commute; the product in
\eqref{eq:m3_interval} therefore collapses to a rotation by the
\emph{accumulated angle}
$\sum_{k=s+1}^{t}\Delta_k\theta_i$ in plane $i$, independent of the
order in which the steps occurred. Mamba-3 thus replaces elapsed time
by a learned, data-dependent measure of elapsed time, and the interval
enters as a scalar summary per plane. The boundary rotations of
\S\ref{sec:tstate} are dense on $\mathbb{R}^{N}$ and do not commute,
so $\Zstate_{c',c}$ depends on the order of the intervening chunks and
not only on an accumulated quantity (\S\ref{app:composition}).

The two therefore answer different questions about an interval.
Mamba-3's factor is a rotation by an angle that grows with the steps
crossed, so what it tells the read-out is \emph{how far back} a frame
was written, in a clock the model has learned rather than in raw
frames; this is why the construction doubles as a relative position
encoding. Ours cannot be reduced to any such quantity: two intervals
of identical length give different $\Zstate$ if the chunks between
them differed, so what the read-out receives is \emph{what came in
between}. The distinction matters wherever two moments are
visually similar but close in time, since their separation is then
small and is in any case already carried by the decay.

\paragraph{Granularity and cost.} Mamba-3 rotates at every step, with
one angle vector per head; the rotation is absorbed into
the projections and the scan itself is untouched, so the cost is that
of a rotary embedding. State regramming rotates once per chunk, with an
$N\times N$ orthogonal matrix predicted from that chunk's content, and
its cost is amortized over $\Cchunk$ frames
(Appendix~\ref{app:impl}). Predicting a full rotation this way is
affordable only because it happens at chunk rate; predicting one per
step would not be.

\paragraph{A separate axis: the discretization scheme.} Mamba-3 also
replaces the zero-order hold with an exponential-trapezoidal rule,
averaging the input contributions at the two ends of an interval
before applying the decay, which reduces approximation error when the
step is large. This is orthogonal to $\lambda$ in the sense that
matters here: it changes how accurately a given interval is
integrated, uniformly along the sequence, and carries no notion of
where an event boundary falls. What $\lambda$ changes is the length of
the interval itself, at frames the task identifies. The two are in
fact complementary, since $\lambda$ enlarges exactly the steps whose
zero-order-hold error a trapezoidal rule would reduce.

\paragraph{Sites.} The two mechanisms operate on different parts of the
recurrence. Mamba-3 applies rotations to the projections, whereas
state regramming applies them to the state passed across chunk
boundaries, leaving $B$ and $C$ unchanged. Because they act on
different components, the two may be complementary, although we do not
evaluate this combination here.

\section{Implementation details}
\label{app:impl}

\paragraph{Cross-clip carry.} On the slow path the SSM hidden state at
the start of a clip is initialized from the previous clip's final
state, and the depthwise-convolution buffer is seeded with that clip's
last $d_\text{conv}-1$ frames, so the forward pass is identical to
running on the concatenated stream while gradients remain truncated at
the clip boundary.

\paragraph{Fusion.} The fast path's input projection acts on the
channel-wise concatenation
$[\,x_\text{fast}^\text{conv}\Vert_\text{ch} y_\text{slow}\,]$, which
is what makes $(\Delta^\text{fast}, B^\text{fast}, C^\text{fast})$
depend on the slow summary. The two path outputs are combined as
$\text{out}=\text{Linear}(\text{RMSNorm}((y_\text{slow}+y_\text{fast})\cdot\text{SiLU}(z)))$,
where the gating stream $z$ is emitted by the fast input projection
alongside $x_\text{fast}$. Summing rather than concatenating keeps the
SSM inner dimension at the canonical SSD value and avoids doubling the
output projection.

\paragraph{Rotation cost.} Both MLPs of \eqref{eq:uvtheta} hold
independent weights per head and
are evaluated with a single batched einsum. Per head and per chunk,
the Cayley map of \eqref{eq:cayley} costs one $N\times N$ solve and
the state update one $N\times N$ multiply; both are amortized over
$\Cchunk$ frames, so the per-frame cost is unchanged in order and the
added parameters are $O(HNr)$. The same operation, with separate
parameters, runs on both paths.

\paragraph{Visual backbone.} \citet{rivoir2024bnpitfalls} showed that
BatchNorm impedes end-to-end training in surgical phase recognition
and used ConvNeXt~\cite{liu2022convnext} as the visual backbone
instead, a choice DACAT~\cite{yang2024dacat} also follows. We train
end-to-end with the same backbone, an ImageNet-pretrained
ConvNeXt-Tiny with the bottom two stages frozen and the top two
fine-tuned jointly with the temporal model.

\paragraph{Training.} Frames are resized to $250\times250$ and randomly cropped to
$224\times224$ with standard augmentation. We train end-to-end with
AdamW on clips of $L{=}256$ frames, chunk size $\Cchunk{=}64$,
$K{=}4$ blocks, state dimension $N{=}64$, and rotation rank $r{=}16$;
the loss weights of \S\ref{sec:loss} are $w_\text{sm}{=}0.15$ and
$w_\text{trans}{=}0.5$, and the transition target uses
$\sigma_L{=}2$ and $\sigma_R{=}12$ frames. Sensitivity to the four core architectural
choices is reported in Appendix~\ref{app:hparam}.

\section{Hyperparameter sensitivity}
\label{app:hparam}

\begin{figure}[t]
    \centering
    \includegraphics[width= 0.8\linewidth]{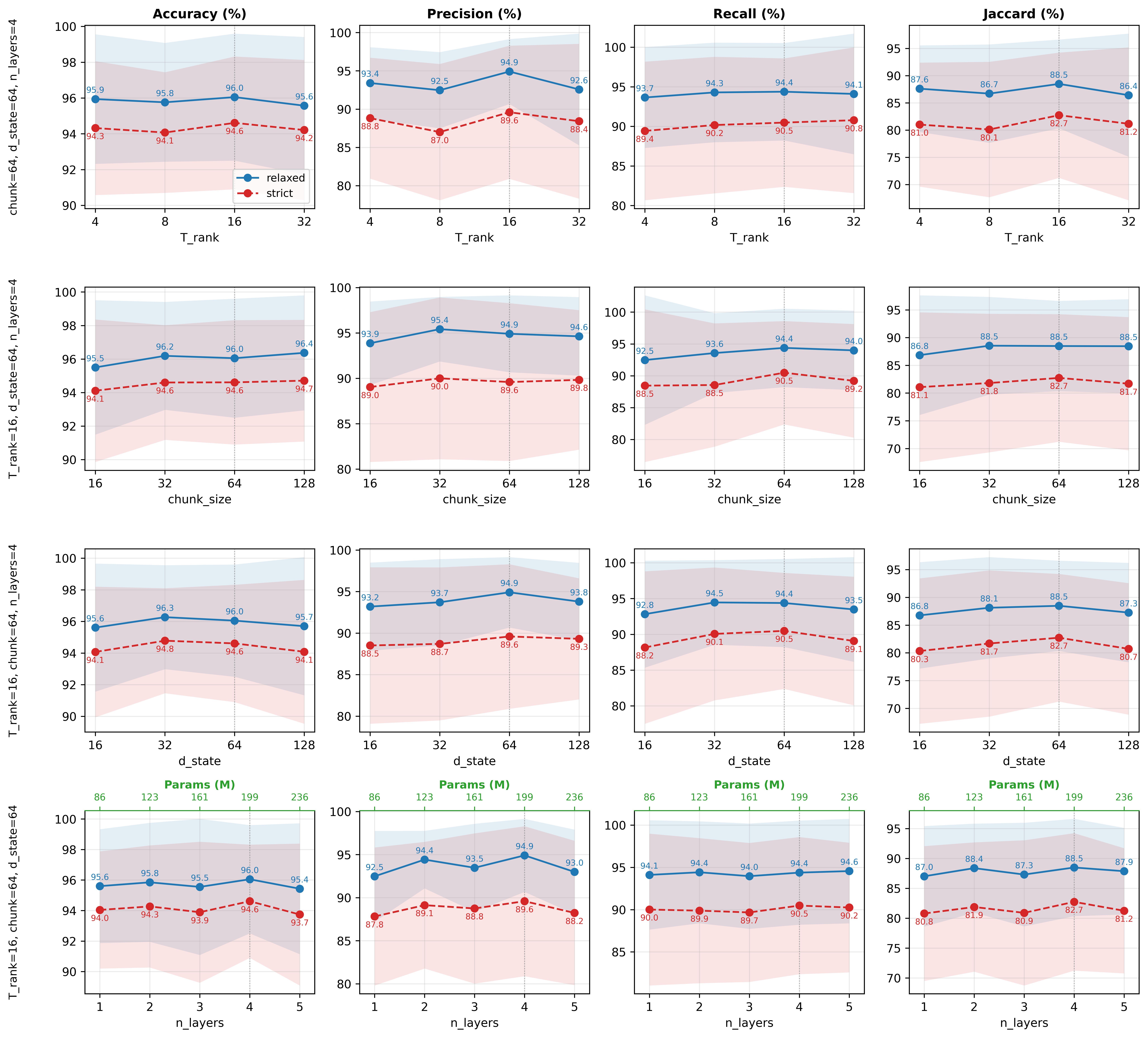}
    \caption{Hyperparameter sensitivity on Cholec80. Each of the rotation rank $r$, chunk size $\Cchunk$, state dimension $N$, and block count $K$ is swept with the other three fixed at the default ($r{=}16, \Cchunk{=}64, N{=}64, K{=}4$; dotted lines); the $K$ sweep also reports parameter count on its top axis. Solid blue/dashed red are the relaxed/strict protocols; shading is one std over the 40 test videos. Strict-accuracy means vary within $1.0$\,pp, well inside the inter-video std, indicating robustness to these choices.}
    \label{fig:hparam}
\end{figure}

Each of the four core hyperparameters (rotation rank $r$, chunk size
$\Cchunk$, state dimension $N$, and block count $K$) is swept on
Cholec80 with the other three held at their defaults ($r{=}16$,
$\Cchunk{=}64$, $N{=}64$, $K{=}4$), under both the relaxed and strict
protocols (Fig.~\ref{fig:hparam}). Every sweep stays within
$1.0$\,pp of the default in strict accuracy, well inside the
inter-video standard deviation, so none of these choices is
load-bearing for the reported results. The rank sweep flattens beyond
$r{=}16$, which says the rotation needs enough planes to re-express
the carried state but not a full-rank map, and this is what makes the
low-rank Cayley parameterization (\S\ref{sec:tstate}) affordable at
chunk granularity.

\section{MQAR protocol and full results}
\label{app:mqar}

\begin{figure}[h]
\centering
\includegraphics[width=\textwidth]{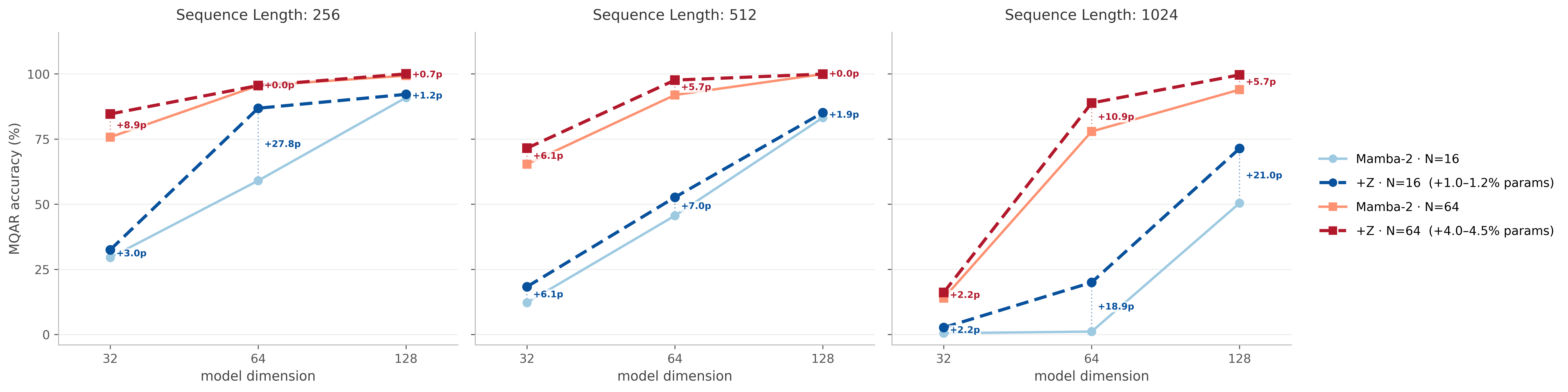}
\caption{MQAR at sequence lengths $256$, $512$ and $1024$ ($T/4$
key--value pairs each), for state sizes $N\in\{16,64\}$ and model
dimensions $\{32,64,128\}$. Labels give the gain from adding $Z$ at
matched $N$. Every configuration improves or is unchanged, the
gain going to zero only where the baseline is already saturated
($N{=}64$ at $d_{\text{model}}{=}128$ for $T\in\{256,512\}$).}
\label{fig:mqar_all}
\end{figure}

We reuse the setting of \citet{dao2024transformers} without
modification. At sequence length $T$ the sequence carries $T/4$
key--value pairs, so $256$ pairs at $T{=}1024$, and the remaining
positions are filled with random distractor tokens drawn from a
vocabulary of $8192$, the harder variant in which non-key/value
positions are randomized rather than left empty. Training cycles
through datasets of $(T/32, T/16, T/8, T/4)$ pairs as a curriculum.
Models are two layers deep, and the learning rate is swept separately
for every configuration, ours and the baseline alike, so that no
variant is reported at another's optimum. The only change we make is
the boundary rotation of \S\ref{sec:tstate}. The chunk size is set to
$T/4$ at every sequence length, so each run crosses four boundaries
and accumulates the same number of rotations regardless of $T$. The
intensity $\lambda$ is not used, as the task provides no transition
labels to supervise it.

Figure~\ref{fig:mqar_all} gives all three sequence lengths. The shape
of the gain is not the same at each. At $T{=}256$ the largest effect
is at $N{=}16$, $d_{\text{model}}{=}64$ ($+27.8$\,pp), at $T{=}1024$
it moves to $d_{\text{model}}{=}128$ ($+21.0$\,pp), and the $N{=}64$
curves saturate earlier at the shorter lengths. Two things are common
across all three. The rotation never hurts, and it helps most where
the pairs to be stored outnumber the directions available to hold
them.

\section{Additional benchmark results}
\label{app:morebench}

Metrics follow standard SPR conventions~\cite{jin2018svrcnet,
funke2023metrics, huang2025mttrnet}. Write $TP_p^v$, $FP_p^v$ and
$FN_p^v$ for the true-positive, false-positive and false-negative
frame counts of phase $p$ in video $v$, with $P$ phases and $V$ test
videos. Accuracy is computed per video and averaged over videos,
\begin{equation}
\text{Accuracy}=\frac{1}{V}\sum_{v}\frac{\sum_p TP_p^v}{T^v},
\end{equation}
with $T^v$ the number of frames in video $v$. Precision, recall and
Jaccard are computed per phase and then averaged across phases within
a video and across videos,
\begin{equation}
\text{Precision}=\frac{1}{PV}\sum_{v}\sum_{p}\frac{TP_p^v}{TP_p^v+FP_p^v},
\qquad
\text{Recall}=\frac{1}{PV}\sum_{v}\sum_{p}\frac{TP_p^v}{TP_p^v+FN_p^v},
\end{equation}
\begin{equation}
\text{Jaccard}=\frac{1}{PV}\sum_{v}\sum_{p} \frac{TP_p^v}{TP_p^v+FP_p^v+FN_p^v}.
\end{equation}

This appendix reports the four benchmarks not covered in
\S\ref{sec:sota}. Table~\ref{tab:datasets} first lists all seven
datasets used in the paper with their class and video counts and
train/val/test splits.

\begin{table}[htbp]
\centering
\caption{Details of the datasets used in our experiments. All splits
follow prior work~\cite{rivoir2024bnpitfalls, jin2021tmrnet}.}
\label{tab:datasets}
\small
\setlength{\tabcolsep}{4pt}
\begin{tabular}{lccc}
\toprule
Dataset & \# classes & \# videos & train : val : test \\
\midrule
Cholec80 \cite{twinanda2016endonet}         & 7  & 80  & 32 : 8 : 40 \\
M2CAI16 \cite{twinanda2016m2cai}            & 8  & 41  & 20 : 7 : 14 \\
Cataract-101 \cite{schoeffmann2018cataracts} & 10 & 101 & 63 : 10 : 28 \\
AutoLaparo \cite{wang2022autolaparo}        & 7  & 21  & 10 : 4 : 7 \\
HeiChole \cite{wagner2023heichole}          & 7  & 24  & 12 : 6 : 6 \\
Heidelberg \cite{maierhein2021heico}        & 14 & 30  & 18 : 6 : 6 \\
GraSP \cite{ayobi2024grasp}                 & 11 & 13  & 6 : 2 : 5 \\
\bottomrule
\end{tabular}
\end{table}

\paragraph{Cataract-101 and HeiChole.}
Table~\ref{tab:cataract_heichole} compares both under their standard
protocols. Cataract-101 is a saturated benchmark, with the top methods within a
point of each other in accuracy, so the gain concentrates in
Jaccard, the metric that still separates them. HeiChole is the
opposite regime, with 24 videos, a different institution, and
considerable domain shift, and it produces the largest margin we observe anywhere
($+5.8$\,pp Jaccard). The pairing is informative because the two
datasets differ mainly in size and shift rather than in task
structure, and the advantage grows in the direction of less data.

\begin{table}[htbp]
\centering
\caption{State-of-the-art comparison on Cataract-101 and HeiChole, each model trained per dataset. Best per block in bold; ``--'' not reported. Both follow their standard protocols.}
\label{tab:cataract_heichole}
\footnotesize
\setlength{\tabcolsep}{3pt}
\begin{tabular}{lcccc}
\toprule
Method & Accuracy & Precision & Recall & Jaccard \\
\midrule
\multicolumn{5}{l}{\textit{Cataract-101}}\\
\midrule
Qi et al. \cite{qi2019cataract}      & 88.1 & -- & -- & -- \\
He et al. \cite{he2022activity}      & 94.5 & 93.1 & 91.6 & -- \\
 MTRCNet-CL \cite{jin2020mtrcnet}       & 94.7 & 92.3 & 91.8 & 85.7 \\
RCNeSt \cite{rcnest}                 & 95.4 & 92.6 & 92.3 & 85.8 \\
CB-RCNeSt \cite{rcnest}              & 96.4 & 94.9 & 94.7 & 90.2 \\
MTTR-Net \cite{huang2025mttrnet}     & 96.9 \(\pm\) 2.7 & 95.7 \(\pm\) 3.2 & 96.2 \(\pm\) 2.2 & 92.1 \(\pm\) 4.8 \\
\textbf{SurgicalMamba (Ours)}        & \textbf{96.9 \(\pm\) 3.1} & \textbf{96.2 \(\pm\) 2.5} & \textbf{96.7 \(\pm\) 2.2} & \textbf{93.2 \(\pm\) 3.7} \\
\midrule
\multicolumn{5}{l}{\textit{HeiChole}}\\
\midrule
ResNet50 \cite{he2016resnet}         & 68.5 \(\pm\) 12.3 & 66.0 & 58.0 & 44.7 \\
SV-RCNet \cite{jin2018svrcnet}       & 70.2 \(\pm\) 11.6 & 67.5 & 59.6 & 45.2 \\
TeCNO \cite{czempiel2020tecno}       & 78.3 \(\pm\) 8.8 & 79.7 & 69.9 & 58.3 \\
Trans-SVNet \cite{gao2021transsvnet} & 78.0 \(\pm\) 9.6 & 77.9 & 68.0 & 56.4 \\
MTTR-Net \cite{huang2025mttrnet}     & 80.1 \(\pm\) 10.9 & 80.3 & 75.8 & 64.5 \\
\textbf{SurgicalMamba (Ours)}        & \textbf{86.4 \(\pm\) 8.7} & \textbf{82.0} & \textbf{78.0} & \textbf{70.3} \\
\bottomrule
\end{tabular}
\end{table}

\paragraph{Heidelberg (HeiCo).}
Table~\ref{tab:heico} reports the 14-phase, 30-video multi-procedure
benchmark, where SurgicalMamba improves Jaccard by $+3.3$\,pp. With
twice the phase count of Cholec80 and fewer videos per phase, this is
the setting in which a recurrence must hold context across many short,
easily confused segments, consistent with the boundary behavior
documented in \S\ref{sec:qualitative}.

\begin{table}[htbp]
\centering
\caption{Comparison on the Heidelberg (HeiCo) dataset. Best in bold.}
\label{tab:heico}
\small
\setlength{\tabcolsep}{6pt}
\begin{tabular}{lcc}
\toprule
Method & Accuracy & Jaccard \\
\midrule
Dylan et al. \cite{dylan2017heico}          & 21 & 8 \\
Andrei et al. \cite{andrei2017heico}        & 57 & 25 \\
Robin et al. \cite{robin2017heico}          & 60 & 38 \\
Sebastian et al. \cite{bodenstedt2017heico} & 61 & 40 \\
MTTR-Net \cite{huang2025mttrnet}            & 70 \(\pm\) 14 & 44 \(\pm\) 20 \\
\textbf{SurgicalMamba (Ours)}               & \textbf{72.1 \(\pm\) 15.1} & \textbf{47.3 \(\pm\) 18.1} \\
\bottomrule
\end{tabular}
\end{table}

\paragraph{GraSP.}
Table~\ref{tab:grasp} reports the prostatectomy benchmark under its
official phase-level mean Average Precision. Here SurgicalMamba does
not improve the mean, matching MTTR-Net at $77.7\%$ mAP. What it
does change is the spread, reducing the per-video standard deviation
from $13.8$ to $12.4$ over the five test videos. With five videos this
is a weak signal on its own, but it is the same direction as the
per-video standard deviations on Cholec80 (Table~\ref{tab:main})
and points to consistency across videos rather than a higher ceiling
on any one of them.

\begin{table}[htbp]
\centering
\caption{Comparison on the GraSP dataset using the official phase-level mean Average Precision (mAP). Best in bold.}
\label{tab:grasp}
\small
\setlength{\tabcolsep}{6pt}
\begin{tabular}{lc}
\toprule
Method & mAP \\
\midrule
SlowFast \cite{feichtenhofer2019slowfast}  & 70.7 \\
TAPIR \cite{ayobi2024grasp}                 & 74.6 \\
TAPIS-VST \cite{ayobi2024grasp}             & 70.9 \\
TAPIS \cite{ayobi2024grasp}                 & 76.1 \\
MTTR-Net \cite{huang2025mttrnet}            & 77.7 \(\pm\) 13.8 \\
\textbf{SurgicalMamba (Ours)}                & \textbf{77.7 \(\pm\) 12.4} \\
\bottomrule
\end{tabular}
\end{table}

Taken together, the four results follow the pattern stated in
\S\ref{sec:sota}. The margins are widest on the smallest and most
domain-shifted sets, and narrowest where the benchmark is saturated or
the test split is small.

\section{Additional qualitative views}
\label{app:qual}

\begin{figure}[htbp]
\centering
\begin{minipage}[t]{0.49\linewidth}
\centering
\makebox[\linewidth][l]{\footnotesize\textbf{(a)}}\par
\includegraphics[width=\linewidth]{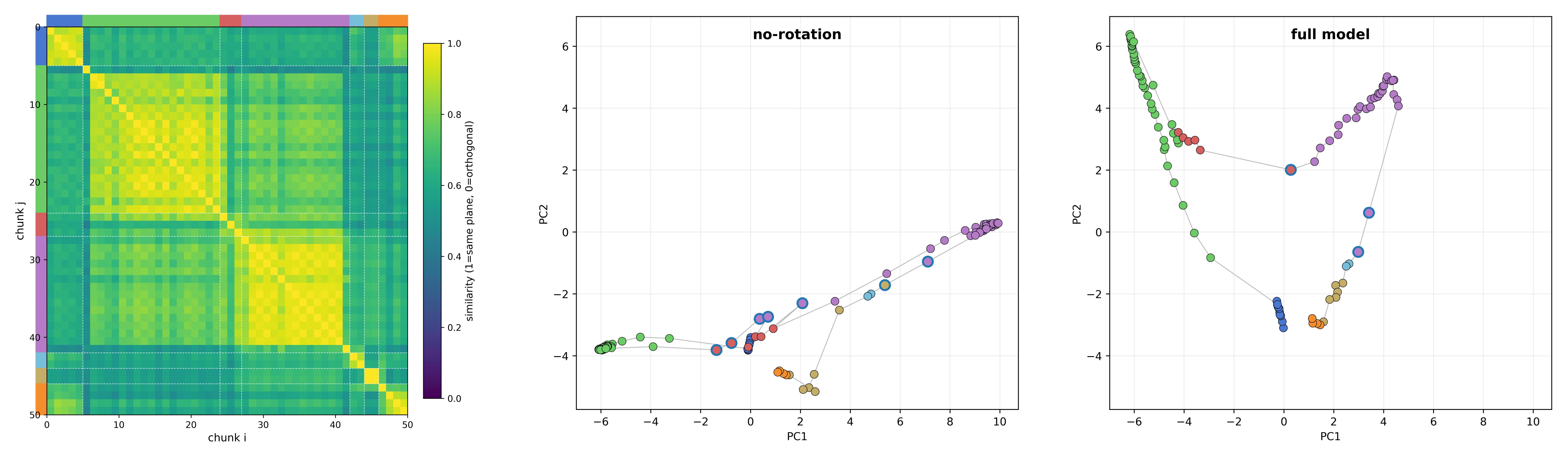}
\end{minipage}\hfill
\begin{minipage}[t]{0.49\linewidth}
\centering
\makebox[\linewidth][l]{\footnotesize\textbf{(b)}}\par
\includegraphics[width=\linewidth]{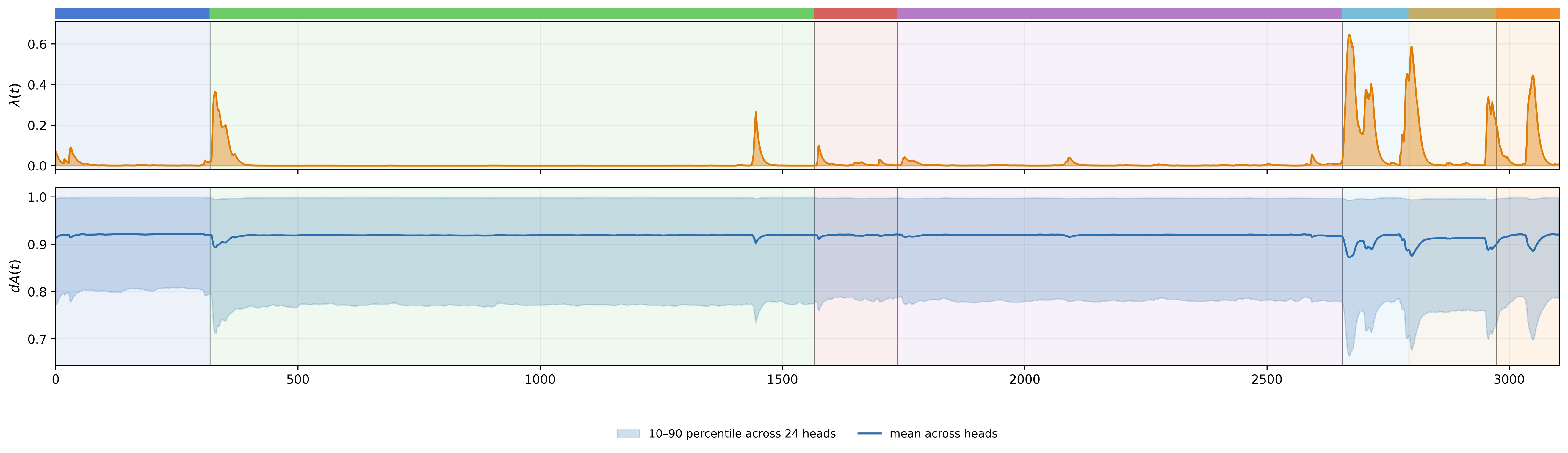}
\end{minipage}
\caption{Internal views on Cholec80 video~41. \textbf{(a)} Rotation
planes: chunk-to-chunk cosine similarity (left) and the slow-path
hidden state under a shared PCA projection, with and without $Z$
(middle, right). \textbf{(b)} Intensity $\lambda(t)$ and the effective
decay $\mathrm dA(t)$, full model versus the no-$\lambda$ ablation.}
\label{fig:qual_app}
\end{figure}

\S\ref{sec:qualitative} reports the predicted phase sequence for
Cholec80 video~41. This appendix gives the two internal views for the
same video, in Fig.~\ref{fig:qual_app}.

\paragraph{Intensity and effective decay.}
In Fig.~\ref{fig:qual_app}(b), $\lambda$ stays near zero through the long
\emph{CalotDiss}/\emph{GBDiss} phases and concentrates at boundaries,
most densely at the late three-phase sequence. The effective decay
follows, holding a plateau near $0.92$ within phases and dipping
exactly where $\lambda$ peaks. The dip itself is an identity of the parameterization
(\S\ref{sec:slowpath}); what the panel documents is that the
supervised signal fires sparsely and precisely at workflow changes.
Each boundary thus triggers a fast state turnover while the plateau
keeps within-phase predictions stable. The no-$\lambda$ variant lacks
the dips, flickers (Fig.~\ref{fig:qual}), and inflates the
strict-accuracy standard deviation ($3.71\to6.05$,
Table~\ref{tab:component_ablation}).

\paragraph{Rotation planes encode phase-aligned structure.}
The chunk-to-chunk cosine similarity of the rotation planes
(Fig.~\ref{fig:qual_app}a, left) is high within a phase and drops across
boundaries, so the planes inherit phase structure although no loss is
applied to $Z$ directly. The network as a whole is trained with phase
labels, but nothing constrains the rotation, and what it settles on is
a basis shared within a phase and separable sub-spaces across phases.
The state geometry agrees. Under a common PCA projection
(Fig.~\ref{fig:qual_app}a, middle and right) the full model spreads
phase-colored arcs over both components, while the no-rotation state
collapses to a one-dimensional band and accumulates more
misclassifications.

Two qualifications apply to that comparison. PC1 and PC2 are
variance-ordered projections rather than channels, so the panel does
not identify which directions rotate. And the no-rotation state still
varies meaningfully along PC1, so what collapses is the
\emph{relative} use of the available dimensions and not the absolute
information content. Read that way, the comparison matches the
prediction of \S\ref{sec:tstate}. A carry that cannot re-arrange
stored content occupies fewer effective directions, and restoring that
ability widens the portion of state space the recurrence uses.

\end{document}